\documentclass[10pt,journal,compsoc]{IEEEtran}
\ifCLASSOPTIONcompsoc
  \usepackage[nocompress]{cite}
\else
  \usepackage{cite}
\fi
\ifCLASSINFOpdf
  \usepackage[pdftex]{graphicx}
  \graphicspath{{../figs/}}
  \DeclareGraphicsExtensions{.pdf,.jpeg,.png}
\else
\fi

\usepackage{stfloats}
\usepackage{amsmath}
\usepackage{amssymb}
\usepackage{dsfont}
\usepackage{pgfplots}
\usepackage{xcolor}
\usepackage{lipsum}
\usepackage{graphicx}
\usepackage{caption}
\usepackage{subcaption}
\usepackage{algorithm}
\usepackage{algorithmic}
\usepackage{placeins}
\usepackage{hyperref}
\usepackage{ragged2e}

\usetikzlibrary{automata,positioning}

\floatname{algorithm}{Procedure}

\usepackage{todonotes}

\newcommand{\BlackBox}{\rule{1.5ex}{1.5ex}}  
\newenvironment{proof}{\par\noindent{\bf Proof\ }}{\hfill\BlackBox\\[2mm]}
 
\newtheorem{theorem}{Theorem}
\newtheorem{lemma}[theorem]{Lemma}

\newtheorem{corollary}[theorem]{Corollary}
\newtheorem{definition}[theorem]{Definition}

\newcommand{\latinphrase}[1]{\textit{#1}}
\newcommand{\etal}{\latinphrase{et~al. }}

\begin{document}
%
\title{Deep Private-Feature Extraction}
%
%
%
%

\author{Seyed~Ali~Osia,~Ali~Taheri,~Ali~Shahin~Shamsabadi,~Kleomenis~Katevas,~Hamed~Haddadi,~Hamid~R.~Rabiee%
\IEEEcompsocitemizethanks{\IEEEcompsocthanksitem Seyed~Ali~Osia, Ali~Taheri and Hamid~R.~Rabiee are with the Advanced ICT Innovation Center, Department of Computer Engineering, Sharif University of Technology, Iran.
\IEEEcompsocthanksitem Ali~Shahin~Shamsabadi and Kleomenis~Katevas are with the School of Electronic Engineering and Computer Science, Queen Mary University of London. 
\IEEEcompsocthanksitem Hamed~Haddadi is with the Faculty of Engineering, Imperial College London.}
}

\IEEEtitleabstractindextext{%
\begin{abstract}
\justifying
We present and evaluate \emph{Deep Private-Feature Extractor (DPFE)}, a deep model which is trained and evaluated based on information theoretic constraints. Using the selective exchange of information between a user's device and a service provider, \emph{DPFE} enables the user to prevent certain sensitive information from being shared with a service provider, while allowing them to extract approved information using their model. We introduce and utilize the \emph{log-rank} privacy, a novel measure to assess the effectiveness of \emph{DPFE} in removing sensitive information and compare different models based on their accuracy-privacy tradeoff. We then implement and evaluate the performance of \emph{DPFE} on smartphones to understand its complexity, resource demands, and efficiency tradeoffs. Our results on benchmark image datasets demonstrate that under moderate resource utilization, \emph{DPFE} can achieve high accuracy for primary tasks while preserving the privacy of sensitive information.

\end{abstract}

}

\maketitle

\IEEEdisplaynontitleabstractindextext

%
\IEEEpeerreviewmaketitle

\IEEEraisesectionheading{\section{Introduction}\label{sec:introduction}}

\IEEEPARstart{T}{he} increasing collection of personal data generated by, or inferred from, our browsing habits, wearable devices, and smartphones, alongside the emergence of the data from the Internet of Things (IoT) devices are fueling many new classes of applications and services. These include healthcare and wellbeing apps, financial management services, personalized content recommendations, and social networking tools. Many of these systems and apps rely on data sensing and collection at the user side, and uploading the data to the cloud for consequent analysis. 

While many of the data-driven services and apps are potentially beneficial, the underlying unvetted and opaque data collection and aggregation protocols can cause excessive resource utilization (i.e., bandwidth and energy)~\cite{vallina2012breaking}, and more importantly data security threats and privacy risks~\cite{acquisti2015privacy}. Collection and processing of private information on the cloud introduces a number of challenges and tradeoffs, especially when scalability of data collection and uploading practice are taken into consideration. The data is often fed into machine learning models for extracting insights and features of commercial interest, where the information is exposed to data brokers and service providers. While certain features of the data can be of interest for specific applications (e.g., location-based services, or mobile health applications), the presence of additional information in the data can lead to unintended subsequent privacy leakages~\cite{haris2014privacy, haddadi2014quantified}. Current solutions to this problem, such as cryptography~\cite{garcia2010privacy, fontaine2007survey}, complete data isolation and local processing~\cite{garcia2015edge} are not efficient for big data and techniques relying on deep learning~\cite{osia2017}. In today's data-driven ecosystem, these privacy challenges are an inherent side effect of many big data and machine learning applications.

  
In this paper, we focus on providing privacy at the first step of this ecosystem: the exchange of acquired user data between the end user and a service provider. We propose a novel solution based on a compromise between scalability and privacy. The proposed framework is based on the idea that when preparing data for subsequent analysis by service provider, the end user does not need to hide all the information by means of cryptographic methods, which can be resource-hungry or overly complex for the end-user device. Instead, it might suffice to remove the sensitive parts of the information (e.g., identity features in a face image), while at the same time preserving the necessary information for further analysis. This is also the case in many surveillance applications where a central node is required to process user data that may be sensitive in some aspects. 

\begin{figure}[t!]	
	\centering
	\includegraphics[width=\columnwidth]{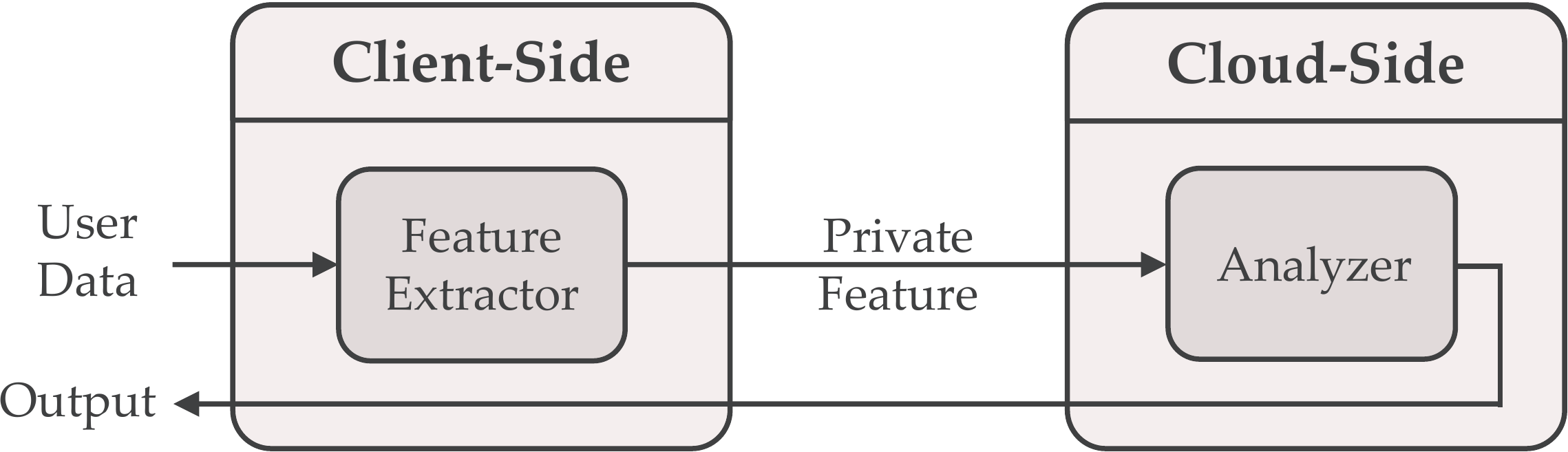}
	\caption{The proposed hybrid framework for user-cloud collaboration.}
	\label{fig:hybrid}
\end{figure}

The proposed hybrid framework in which the user and cloud collaborate to analyze the raw user data in a private and efficient manner is depicted in figure~\ref{fig:hybrid}. Our work relies on the assumption that the service provider releases a publicly verifiable feature extractor module based on an initial training set. The user then performs a minimalistic analysis and extracts a \emph{private-feature} from the data and sends it to the service provider (i.e., \emph{the cloud}) for subsequent analysis. The private-feature is then analyzed in the cloud and the result yields back to the user. The fundamental challenge in using this framework is the design of the feature extractor module that removes sensitive information properly, and on the other hand does not impact scalability by imposing heavy computational requirements on the user's device.

In the rest of this paper, we first discuss the privacy issues in different aspects of machine learning and consider the problem of ``user data privacy in interaction with cloud services" as the main purpose of this paper. To design the feature extractor module, we express our privacy preservation concerns in an optimization problem based on mutual information and relax it to be addressable by deep learning. We then present the Deep Private-Feature Extractor (\emph{DPFE}), a tool for solving the aforementioned relaxed problem. We then propose a new privacy measure, the \emph{log-rank} privacy, to verify the proposed feature extractor, measure its privacy, and evaluate the efficiency of the model in removing sensitive information. The log-rank privacy can be interpreted from different perspectives, including entropy, $k$-anonymity and classification error. We evaluate this framework under the facial attribute prediction problem by using face images. In this context, we remove the face identity information while keeping facial attribute information, and analyze the privacy-accuracy performance tradeoff. Finally, we implement different private-feature extractors on mobile phone to compare the performance of different solutions and addressing the scalability concern.

The main contributions of this paper are: (i) Proposing a hybrid user-cloud framework for the user data privacy preservation problem which utilizes a private-feature extractor as its core component; (ii) Designing the private-feature extractor based on information theoretic concepts leading to an optimization problem (Section~\ref{sec:opt}); (iii)  Proposing a deep neural network architecture to solve the optimization problem (Section~\ref{sec:deep}); (iv) Proposing a measure to evaluate privacy and verify the feature extractor module (Section~\ref{sec:privacy}).\footnote{All the code and models for the paper are available on \url{https://github.com/aliosia/DPFE}}

\section{privacy in machine learning}

\begin{figure}[t]
	\centering
	\includegraphics[width=\columnwidth]{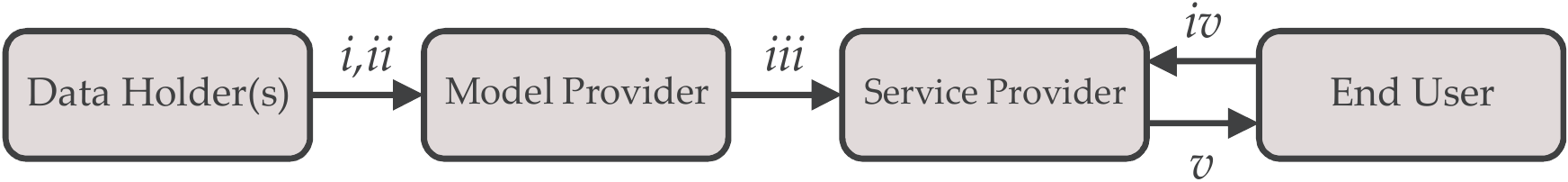} 
	\caption{Privacy concerns may exist when: (i) data holder shares a public dataset: anonymity of individuals are threatened; (ii) data holders participate in a model training procedure with their private data; (iii) a model provider shares a publicly-learned model: the privacy of the individuals' data used for training is at risk; (iv) an end user shares his/her data with the service provider: private information can be revealed to the service provider; (v) a service provider shares query answers with the end user: an attacker can infer the model itself by launching repeated queries.}
	\label{fig:prior}
\end{figure}

Machine learning methods need to analyze sensitive data in many usecases to perform their desired tasks which may violate users' privacy. This fundamental dichotomy has been appearing in different aspects of machine learning, as listed in Figure~\ref{fig:prior}. These concerns can be classified as \emph{public dataset privacy}, \emph{training phase privacy}, \emph{training data privacy}, \emph{model privacy} and \emph{user data privacy}  which are discussed in the rest of this section.

%

\subsection{Public Dataset Privacy}

Training data is the crucial component of each learning system. Collecting and sharing rich datasets for data mining tasks can be highly beneficial to the learning community, although it might come with privacy concerns that make it a double-edged sword. Publishing a dataset that satisfies both parties by preserving the users' privacy and other useful information for data mining tasks, is a challenging problem and has long line of work. Agrawal and Srikant~\cite{agrawal2000} were some of the first to address the privacy concern in data mining for sharing a generic dataset for learning tasks, in addition to considering users' privacy. They utilized a \emph{randomization} technique, in which by adding noise to data they guaranteed its privacy. The resulting distribution of noisy data might have been different from the original distribution. To reconstruct the original distribution, a recovery method was introduced in the paper and extended by Agrawal \etal in~\cite{agrawal2001}. By utilizing this method, it is possible to train a learning model on reconstructed data with the same distribution as the original data. Many works have followed this trend and extended this idea, however, this approach faces two important obstacles: curse of dimensionality and non-robustness to attacks~\cite{aggarwal2008}, which make it inefficient for high dimensional data with side informations. 

\emph{$k$-anonymity} is another popular option for addressing the problem of anonymous dataset publishing, first introduced by Sweeney~\cite{sweeney2002}. Publishing a health database that contains patient sensitive information is one of the favored instance of $k$-anonymity usages. Assuming all data points have identity documents (IDs) that should be kept private, $k$-anonymity deals with transforming a dataset in a way that, having an individual data features, one cannot infer its ID among at least $k$ identities. Many researches are presented to make a database $k$-anonymous~\cite{fung2005,wang2004,bayardo2005,lefevre2006} and most of them are based on the generalization (e.g. removing the last digit of the patient zip code) or suppression of features (e.g. removing the name). Nevertheless, this approach deals with some important challenges when facing attacks~\cite{aggarwal2008}, although~\cite{machanavajjhala2007}, \cite{li2007} and~\cite{rebollo2010} tried to overcome these challenges. Furthermore, they are only well-suited for structured databases with high level features (e.g. relational databases) which makes them hard to deploy for other type of data (e.g. image and video). Newton~\etal~\cite{newton2005} published a $k$-anonymous image dataset by proposing the $k$-same algorithm. While they build the desired dataset by constructing average images among $k$ identities, their employed models are not reliable today.


\subsection{Training Phase Privacy}

A common problem of centralized learning is the collection of training data, especially when dealing with individual's sensitive data (e.g. health information). People are usually reluctant in sharing data that includes their habits, interests, and geographical positions. The upcoming solution to this problem is federated learning, where data holders keep their data private, while they communicate with a central node in order to train a learning model in a cooperative manner. \cite{shokri2015} tried to address this problem by using distributed stochastic gradient descent (\emph{SGD}), where each party loads the latest parameters, update them using \emph{SGD} and upload the new selected parameters to the central node that holds the global model. While in that case direct leakage of private data can be prevented, the uploaded gradients might still include sensitive information from the training data. Thus, a differentially private algorithm is required for sharing the gradients which is proposed in that work. This approach still has some major problems e.g. loose privacy bound addressed by~\cite{papernot2016} and potential threats by generative adversarial networks addressed by~\cite{hitaj2017}. An alternative solution to this problem could be the use of cryptographic techniques like secure multi-party computation, recently used by~\cite{mohassel2017}. However, these techniques are still not applicable on complex neural networks, due to their low efficiency and accuracy.


\subsection{Training-Data Privacy}

The growing popularity of public learning models raises the concern of privacy of the individuals involved in the training dataset. Differentially private algorithms brought us a rigorous answer to this problem, by providing a method to answer queries from a statistical database, without disclosing individuals' information, as formalized by~\cite{dwork06}. An algorithm is called differentially private if the conditional likelihood ratio of presence and absence of an individual, given the transformed statistic, is close to one. Adding noise to the original statistic is one popular method leading to differential privacy. We can consider a learning model as a complex statistic of its training data which should not reveal information about the individuals. Answering complex queries by combining simple queries is the way various learning models, such as Principal Component Analysis and $k$-means, can be made differentially private (see the surveys by~\cite{dwork2008} and~\cite{ji2014}). Recently differentially private deep models were proposed by~\cite{abadi2016}. The authors in~\cite{papernot2016} introduced a privacy preservation framework by utilizing differential privacy which is not specific to the learning model and possesses a state of the art privacy-accuracy tradeoff.

\subsection{Model Privacy}

Model privacy is the concern of the service provider and deals with keeping the learning model private, while returning the inference results to the user. Throughout these years, less attention has been paid to the model privacy, although some works such as~\cite{tramer2016stealing} studied this problem. In general, an adversary can infer the model parameters by making many queries to the learning model and aggregate the answers. \cite{tramer2016stealing} considered this approach for some basic models, e.g. logistic regression, multilayer perceptron and decision tree.

\subsection{User Data Privacy}

The increasing usage of cloud-based systems has triggered a situation where preserving privacy is a challenging but important task. That is, when the user data and the pre-trained learning model is not accessible from the same place, inevitably user data must be sent to the service provider for further analysis. Usually, cryptographic schemes are prevalent in these situations, where two parties do not trust each other. Focusing on the deep models offered by a cloud service, \cite{gilad2016} introduced this problem and proposed a homomorphic encryption method to execute the inference directly on the encrypted data. Even though this work is an interesting approach to the problem, a number of shortcomings makes it impractical. In fact, approximating a deep neural network with a low degree polynomial function may not be feasible without sacrificing accuracy. Furthermore, the complexity of the encryption is relatively high which makes it inefficient for the real-world online applications. 
An alternative to homomorphic encryption was suggested by \cite{rouhani2017}. They used garbled circuit protocol and address some of the discussed challenges, however they were limited to employing simple neural networks and had very high computational cost. 

In summary, using cryptographic techniques on complex deep neural networks is not feasible yet, while the problem of user data privacy is getting more and more important everyday in the cloud computing era. In this paper we are targeting this challenge and try to address it with a machine learning solution, based on a specific kind of feature extraction model, formulated in the next section.


\section{Problem Formulation}
\label{sec:opt}

In this section, we address the user data privacy challenge in a different manner from encryption-based methods. The key intuition is that for many applications we can remove all of the user's \emph{sensitive} (unauthorized) information while retaining the ability to infer the \emph{primary} (authorized) information. This is as opposed to encryption-based solutions that try to encode all the information such that only authorized users can access it. For instance, we may want to focus on hiding individuals' identities in a video surveillance system, but still allow to count the number of participants. In this scenario, a trivial solution is to censor people's faces in the frames, however this solution fails when the purpose is to measure the facial attributes such as emotion or gender. Henceforth, we address this problem as a \emph{privacy preservation problem} and use the terms \emph{primary} and \emph{sensitive} information as the information needed to be preserved and removed, respectively. Assuming the service provider knows the primary and sensitive random variables, we abstract this concept as an optimization problem by utilizing mutual information (see Appendix~\ref{appendix:A} for information theoretic preliminaries). 

Let $\textbf{x}$ be the input, $\textbf{z}$ the primary, and $\textbf{y}$ the sensitive variables. We would like to extract a feature $\textbf{f}$, by applying a function $\textbf{g}$ on $\textbf{x}$, which is informative about the primary variable and non-informative about the sensitive variable. We refer to the extracted feature as \emph{private-feature}. More specifically, the desired private-feature is obtained through maximizing mutual information between the feature and primary variable $I(\textbf{f}; \textbf{z})$, while minimizing mutual information between the feature and sensitive variable $I(\textbf{f};\textbf{y})$, as follows:


\begin{align*}
\begin{split}
\max_{\textbf{f}}\; & I(\textbf{f};\textbf{z}) - \beta I(\textbf{f};\textbf{y})  \\
s.t. \; & \textbf{f} = g(\textbf{x})
\end{split}
\end{align*}

where $I(A;B)$ represents the mutual information between two random variables $A$ and $B$.

Even though at the first glance it seems that the optimal solution of this problem is to set $\textbf{f}$ equal to the best estimation of $\textbf{z}$, this is not applicable in many real world applications because: (a)~the optimal model which perfectly predicts $\textbf{z}$ can be too complicated, and hence using such a feature extractor in the client-side is impossible; and (b)~the service provider may not share the whole model with the client for some reasons such as copyright issues. Assuming we can accurately estimate $\textbf{f}$ by using a member of family of functions $\mathbb{G}=\{g(x;\theta) | \theta \in \Theta\}$, then the optimization problem becomes:
    
\begin{align}\label{eq:base}
\begin{split}
\max_{\theta}\; & I(\textbf{f};\textbf{z}) - \beta I(\textbf{f};\textbf{y})  \\
s.t. \; & \textbf{f} = g(\textbf{x};\theta)
\end{split}
\end{align}

\begin{figure}[t]
	\centering
	\includegraphics[width=.4 \columnwidth]{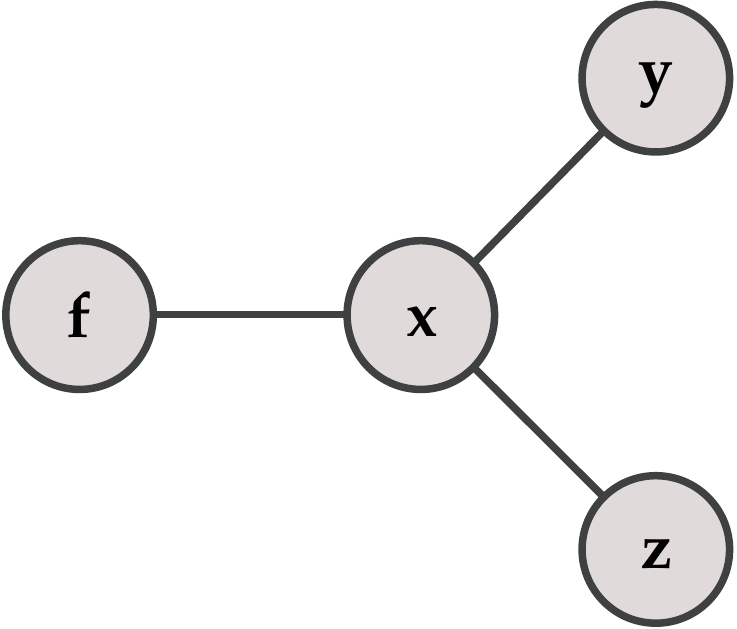} 
	\caption{Private-feature extraction probabilistic graphical model.}
	\label{fig:pgm}
\end{figure}

where, $\textbf{f}$ is a deterministic function of the input variable, parameterized by $\theta$. The graphical model of this problem is shown in Figure~\ref{fig:pgm}.


Optimizing mutual information has been widely used in many information theoretic approaches of machine learning problems. The authors in \cite{bell1995} formulated the \emph{Infomax} and tried to address the problem of unsupervised deterministic invertible feature extraction by maximizing the mutual information between the input and feature. \cite{barber2003} relaxed the limiting invertibility constraint and used a variational approach which leads to the \emph{IM} algorithm for maximizing the mutual information. Recently, \cite{chen2016} used a similar method to maximize the mutual information in generative adversarial networks. These works can be considered as the fundamental works in the problem of unsupervised feature extraction from information theoretic viewpoint, however, since we are utilizing a supervised approach, those methods cannot be applied to our case. Among works considering supervised feature extraction, The \emph{information bottleneck} introduced in~\cite{tishby2000} is the most relevant work. In general, Information bottleneck provides an information theoretic framework for analyzing the supervised feature extraction procedure. Although their optimization problem almost looks similar to ours, there is a fundamental difference between the two approaches. More specifically, they use $I(\textbf{f};\textbf{x})$ instead of $I(\textbf{f};\textbf{y})$, meaning that irrelevant information to $\textbf{z}$ should be removed by minimizing $I(\textbf{f};\textbf{x})$ in the process of feature extraction. Therefore, they can not directly consider the privacy constraints about $\textbf{y}$. Moreover, their optimization problem is solved through an analytical approach that assumes that the joint probability distribution $p(x,z)$, is known. However, in practice this distribution is often unavailable. Although their analytical method is impractical, nevertheless, their framework provides a powerful tool for analysis of the supervised feature extraction methods. 

Similar to the information bottleneck optimization problem, the private-feature extraction problem (Equation~\ref{eq:base}) is non-convex and can not be solved through the known convex optimization algorithms. To overcome this challenge, it is common to bound the optimization problem, and then by using the iterative methods similar to \cite{barber2003} and~\cite{alemi2016}, obtain the desired results. To this end, we first obtain the lower and upper bounds for $I(\textbf{f};\textbf{z})$ and $I(\textbf{f};\textbf{y})$ respectively, and then try to maximize the lower bound of Equation~\ref{eq:base}. Henceforth, we assume $\textbf{y}$ to be a discrete sensitive variable in order to address the classification privacy problem.\\

\textbf{Lower bound for $I(\textbf{f};\textbf{z})$.} We derive a variational lower bound for mutual information by first expressing Lemma~\ref{lem0} and then proving Theorem~\ref{th0}.
\begin{lemma}
	\label{lem0}
	For any arbitrary conditional distribution $q(z|f)$, we have:
	\begin{align}\label{eq:lb0}
	I(\textbf{f};\textbf{z}) \geq \mathds{E}_{\textbf{f},\textbf{z}} \log \frac{q(z|f)}{p(z)}
	\end{align}
	\begin{proof}
		See Appendix~\ref{appendix:B_1}.
	\end{proof}
\end{lemma}

\begin{theorem} \label{th0}
	$\;$ The lower bound $\mathcal{L}$ for $I(\textbf{f};\textbf{z})$ is given by:
	\begin{align}
	\mathcal{L} =  H(\textbf{z}) + \max_{\phi}\; \mathds{E}_{\textbf{f},\textbf{z}} \log q(z|f;\phi)
	\end{align}
	\begin{proof}
		For all members of a parametric family of distributions $\{q(z|f;\phi)|\phi \in \Phi\}$, the right hand side of Equation~\ref{eq:lb0} can be considered as the lower bound for mutual information. The equality happens when $q(z|f)$ is equal to $p(z|f)$. Therefore, if we consider a rich family of distributions for $q$ in which a member can approximate $p(z|f)$ well enough, we can obtain a tight enough lower bound for mutual information by maximizing the right hand side of Equation~\ref{eq:lb0} with respect to $\phi$. By utilizing the definition of Entropy, we obtain $\mathcal{L}$ as the desired lower bound.
	\end{proof}
\end{theorem}

\textbf{Upper bound for $I(\textbf{f};\textbf{y})$.} A two-step procedure can be used to find an upper bound for the mutual information. First, we use the Lemma~\ref{lem1} and Jensen inequality to prove Theorem~\ref{th1}, and obtain $\mathcal{U}_1$ as the primitive upper bound for $I(\textbf{f};\textbf{y})$. Then we use kernel density estimation (\emph{KDE}) (see \cite{bishop2006}) and use Lemma~\ref{lem2} and \ref{lem3} to obtain $\mathcal{U}_2$ as the desired upper bound for $I(\textbf{f};\textbf{y})$ through Theorem~\ref{th2}. 

\begin{lemma}
	\label{lem1}
	$\;$ Assume $\textbf{y}$ is a discrete random variable with $\{y_a|1\le a \le c_y\}$ as its range, then:
	\begin{align*}
	I(\textbf{f};\textbf{y}) = \sum_a p(y_a) \int p(f|y_a) \log \frac{p(f|y_a)}{\sum_b p(y_b)p(f|y_b)}\ df
	\end{align*} 
	\begin{proof}
		By substituting $\frac{p(f,y)}{p(f)p(y)}$ with $\frac{p(f|y)}{p(f)}$ and then $p(f)$ with $\sum_b p(y_b)p(f|y_b)$ in the main formula of $I(\textbf{f};\textbf{y})$, we obtain the desired relation.
	\end{proof}
\end{lemma}

By utilizing Jensen inequality\footnote{see Appendix~\ref{appendix:A}} and manipulating Lemma~\ref{lem1}, we can compute $\mathcal{U}_1$ as a primitive upper bound for mutual information, as follows.
\begin{theorem} \label{th1}
	$\;$ The upper bound $\mathcal{U}_1$ for $I(\textbf{f};\textbf{y})$ is given by:
	\begin{align}
	\mathcal{U}_1 = \sum_a \sum_{b: b\neq a} p(y_a)\ p(y_b)\ D_{kl} \big[ p(f|y_a) \| p(f|y_b) \big]
	\end{align}
	\begin{proof}
		See Appendix~\ref{appendix:B_2}.
	\end{proof}
\end{theorem}


Since computing $\mathcal{U}_1$ by Equation~\ref{th1} is not tractable, we use an approximation technique to obtain the upper bound. By employing the kernel density estimation, we can efficiently estimate $p(f)$ \cite{duong2005}. We then utilize the Silverman's rule of thumb~\cite{silverman1986} and use a Gaussian kernel with the desired diagonal covariance matrix. Next, by normalizing each dimension of the feature space to have zero mean and unit variance, we acquire a symmetric Gaussian kernel with \emph{fixed} covariance matrix, $\sigma I$, where $\sigma$ is a constant depending on the dimensionality of the feature space and the size of the training data. This kind of normalization is a common process in machine learning~\cite{lecun1998} and is impartial of relations among different dimensions including independency and correlation. Finally, conditioning on $\textbf{y}$, for each $y_a$ we can think of $p(f|y_a)$ as a Gaussian Mixture Model (\emph{GMM}) (see \cite{bishop2006}) and use the following lemmas from~\cite{hershey2007} to obtain a reliable upper bound.


\begin{lemma} \cite{hershey2007}
	\label{lem2}
	$\;$ For two multidimensional Gaussian distributions, $p$ and $q$, with $\mu_p$ and $\mu_q$ as their expected values and the same covariance matrix $\sigma I$, we have:
	\begin{align*}
	D_{kl} (p \| q) = \frac{1}{2\sigma} \|\mu_p-\mu_q\|_2^2
	\end{align*} 
\end{lemma}

\begin{lemma} \cite{hershey2007}
	\label{lem3}
	$\;$ For two given GMMs $p=\sum_a \pi_a p_a$  and $q=\sum_b \omega_b q_b$, we have:
	\begin{align*}
	D_{kl} (p \| q) \leq \sum_{a,b} \pi_a \ \omega_b \ D_{kl}(p_a \| q_b)
	\end{align*}
	where for each $a$ and $b$, $p_a$ and $q_b$ are Gaussian distributions forming the mixtures.
\end{lemma}

We can use Theorem~\ref{th1}, Lemma~\ref{lem2} and Lemma~\ref{lem3} to derive the desired upper bound for $I(\textbf{f};\textbf{y})$.

\begin{theorem} \label{th2}
	$\;$Having large training data, the upper bound $\mathcal{U}_2$ for $I(\textbf{f};\textbf{y})$ is given by:
	\begin{align}\label{eq:u2maina}
	\mathcal{U}_2 = \frac{1}{\sigma N^2} \sum_{\substack{(i,j):\\y_i\neq y_j}} \|f_i-f_j\|_2^2 
	\end{align}
	where $f_i$ is the extracted feature from data $x_i$ and its corresponding label $y_i$. The sum is over pairs of points with different $\textbf{y}$ labels.
	\begin{proof}
		See Appendix~\ref{appendix:B_3}.
	\end{proof}
\end{theorem}

In other words, $\mathcal{U}_2$ is an upper bound that is proportional to the average Euclidean distance between each pairs of feature vectors having different $\textbf{y}$ labels. This value is practically hard to estimate, especially when we use \emph{SGD}, and have large number of classes. Therefore, as stated in Theorem~\ref{th3} and Corollary~\ref{corol}, we use an equivalent relation for $\mathcal{U}_2$ which is easier to optimize.

\begin{theorem} \label{th3}
	Constraining the variance of each dimension of feature space to be 1, we have:
	\begin{align}\label{eq:u2main}
	\mathcal{U}_2 = \frac{1}{\sigma N^2}\sum_{\substack{(i,j):\\y_i = y_j}} (c - \|f_i-f_j\|_2^2) 
	\end{align}
	where $c$ is a constant function of feature space dimension and number of training data. 
	\begin{proof}
		See Appendix~\ref{appendix:B_4}.
	\end{proof}
	\begin{corollary} \label{corol}
		We can optimize the right hand side of Equation~\ref{eq:u2main} instead of Equation~\ref{eq:u2maina} fo obtain $\mathcal{U}_2$. 
	\end{corollary}
\end{theorem}

Considering Corollary~\ref{corol} together with Theorem~\ref{th3}, we realize that for a random pair of feature points, we should decrease their distance if the $y$ labels are different, and increase their distance if they are the same. This is very similar to the \emph{Contrastive} loss idea presented in~\cite{hadsell2006} which is a popular loss function for \emph{Siamese} architecture~\cite{chopra2005}. Siamese networks are used for metric learning purposes and tends to form a feature space in which similar points are gathered near each other. This is the opposite of what we aim to achieve: increase the distance of similar points and decrease the distance of dissimilar points. 



By utilizing the suggested lower and upper bounds, we can substitute the original private-feature extraction problem (Equation~\ref{eq:base}) with the following relaxed problem.

\begin{align} \label{eq:relaxed}
\begin{split}
\min_{\theta, \phi}\; 
& \sum_{f_i,z_i} -\log q(z_i|f_i;\phi) \\ 
&  + \frac{\beta}{2\sigma N^2} \Big[ \sum_{\substack{(i,j):\\y_i\neq y_j}} \|f_i-f_j\|_2^2  +  \sum_{\substack{(i,j):\\y_i = y_j}} (c - \|f_i-f_j\|_2^2)  \Big]  \\
s.t.\; & f_i = g(x_i;\theta)
\end{split}
\end{align}

\begin{figure}[t]
	\centering
	\includegraphics[width=.5\columnwidth]{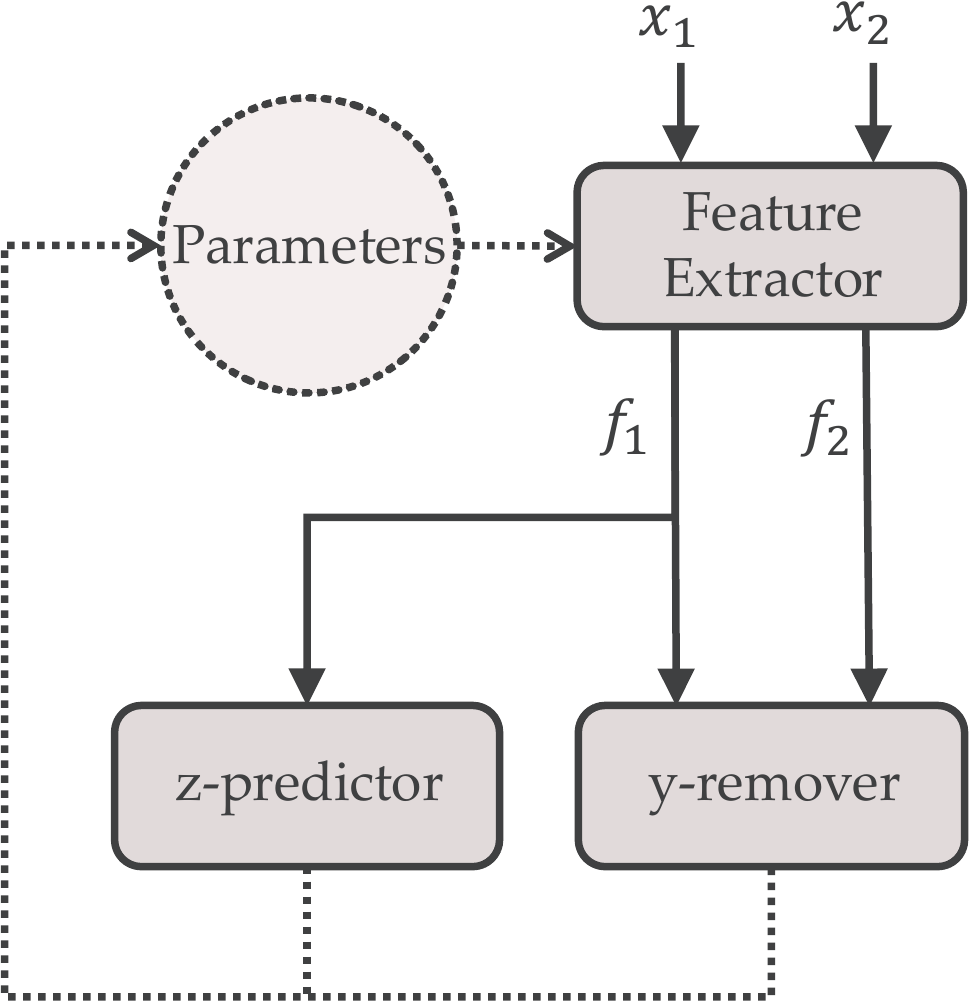} 
	\caption{The private-feature extraction framework. $z$ and $y$ are the primary and sensitive variables, respectively. $x_1$ and $x_2$ are two independent samples, and $f_1$ and $f_2$ are their corresponding features. $\textbf{z}$-predictor just uses $f_1$ to compute the first term of loss function, whereas $\textbf{y}$-remover uses both $f_1$ and $f_2$ to compute the second term of loss function (see Equation~\ref{eq:relaxed}). Solid lines show data flow and dotted lines indicate affection.}
	\label{fig:framework}
\end{figure}

Considering the above equation, we should optimize an objective function that consists of two loss functions: the loss of the primary variable preservation modeled by a classification loss (first term), and the loss of sensitive variable elimination modeled by a contrastive loss (second term). Thus, the general training framework of the private-feature extractor contains three main modules: feature extractor, primary variable predictor and sensitive variable remover, as shown in Figure~\ref{fig:framework}. Note that according to the second term of Equation~\ref{eq:relaxed}, the loss function of removing sensitive variable is defined on the pairs of samples, and as a result the $\textbf{y}$-remover module also operates on pairs of features.

We propose a general deep model along with \emph{SGD}-based optimizers to solve the optimization problem in Equation~\ref{eq:relaxed}, as explained in the next section.


\section{Deep Architecture} \label{sec:deep}

By utilizing the latest breakthroughs in the area of deep neural networks, we can practically find good local optimums of non-convex objective functions through \emph{SGD}~based algorithms, and accurately estimate complex non-linear functions. Today, a large portion of state of the art learning models are deep. Therefore, having a general framework for privacy preserving deep inference is necessary. 
In this paper, we focus on image data (in the context of identity v.s. gender, expression, or age recognition), and propose a deep architecture based on CNN (Fig.~\ref{fig:arc}) to optimize the objective function of the relaxed problem (Equation~\ref{eq:relaxed}). It is worth mentioning that the proposed framework can be generalized to other applications and deep architectures (e.g. recurrent neural networks).

We call the proposed \emph{Deep Private-Feature Extractor} architecture; \emph{DPFE}. We consider two consecutive CNNs; one as the feature extractor and the other as the primary variable predictor. A simple strategy for building these modules is the \emph{layer separation mechanism} introduced in~\cite{osia2017_2}. We can also employ batch normalization layer \cite{ioffe2015} and normalize each dimension of the feature space, as stated in Section~\ref{sec:opt}. In the following, we first introduce the layer separation mechanism, and then proceed with dimensionality reduction and noise addition issues that can enhance the preservation of privacy.

\subsection{Layer Separation Mechanism}

\begin{figure}[t]
	\centering
	\includegraphics[width=\columnwidth]{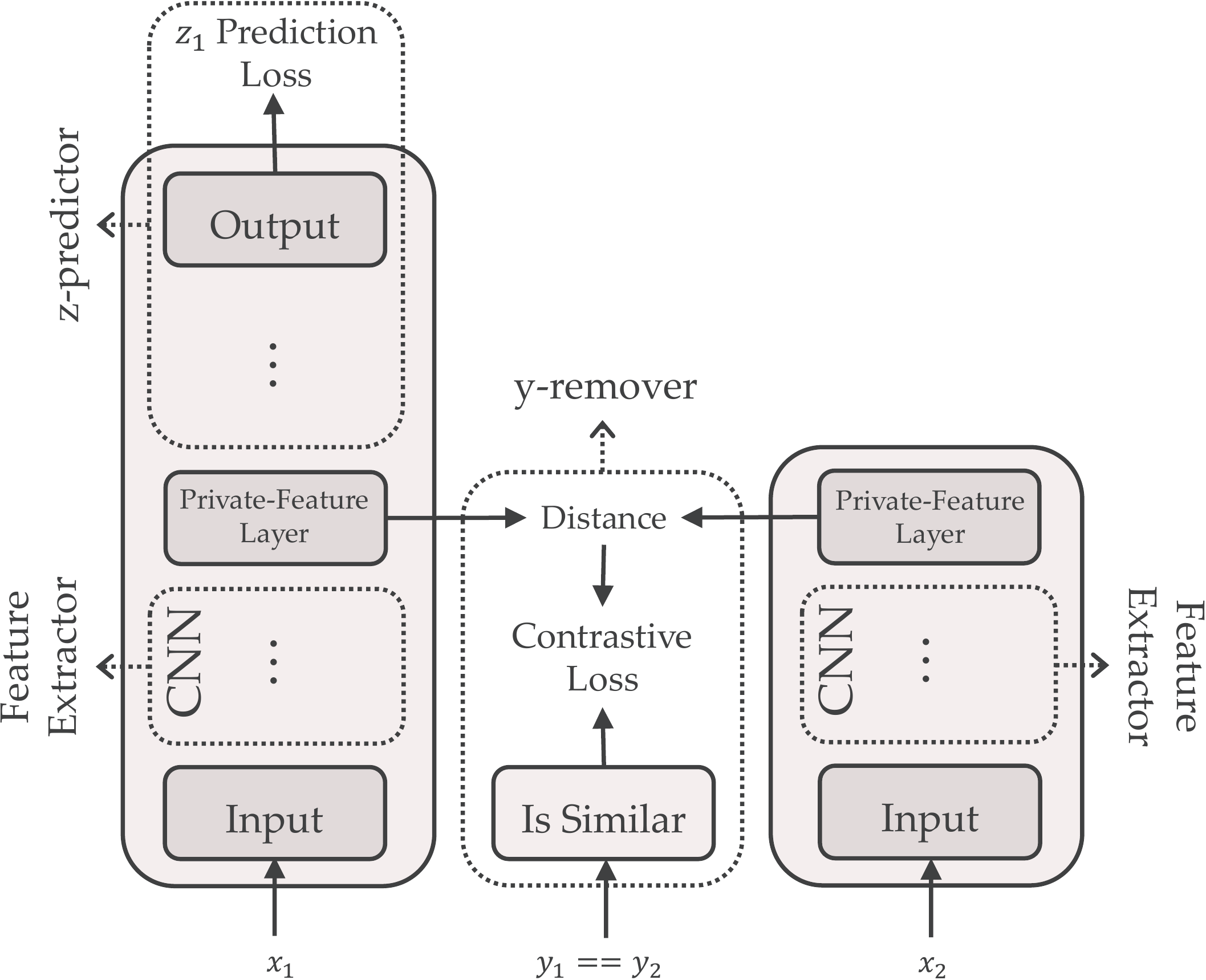} 
	\caption{Deep CNN architecture for private-feature extraction (\emph{DPFE} architecture). $x_1$ and $x_2$ are independent random samples and $y_1$ and $y_2$ are their corresponding sensitive labels. $\textbf{y}$-remover first checks the equality of sensitive labels and then apply the information removal loss function.}
	\label{fig:arc}
\end{figure}

To deploy our framework, we can start from a pre-trained recognizer of primary variable (e.g. a deep gender recognition model), and make it private to the sensitive variable (e.g. identity). In order to do this, we choose the output of an arbitrary intermediate layer of the pre-trained model as the preliminary private-feature and simply partition the layers of the model into two sets: the elementary and the secondary layers that form the feature extractor and the primary variable predictor, respectively. In this way, the same model can be easily fine-tuned by just appending the contrastive loss function and continuing the optimization process, leading to the \emph{private-feature} as the intermediate layer's output. This procedure is shown in Fig.~\ref{fig:arc}.

One may argue that separating the layers of a deep model is sufficient to obtain an ideal private-feature in the intermediate layer due to the nature of deep networks. In general, the higher layers of a deep architecture provide a more abstract representation of the data and drop the irrelevant information including the sensitive information \cite{shwartz2017} and preserve the primary variable ~\cite{yosinski2014}. Therefore, there is no need to fine-tune the model with the suggested \emph{DPFE} architecture. However, this argument can easily be rejected by considering the counter example provided by deep visualization techniques. For example, \cite{dosovitskiy2016} provided a method to reconstruct the input image from intermediate layers of a deep network. Osia et.al. used this method in \cite{osia2017_2} and demonstrated that the original face image can be reconstructed from some intermediate layers of the gender recognition model. Thus, there is no guarantee that the intermediate layers drop the sensitive information (identity in this case).

\subsection{Dimensionality Reduction}

Imagine that the extracted private-feature has a low dimension (in Section~\ref{sec:privacy} we use a 10-dimensional feature space). In this case, we will benefit from the following advantages: 
\begin{itemize}
	\item We can highly decrease the communication cost between user and the service provider, because instead of sending the raw input data to the cloud, the user will only send the private-feature of the input.
	\item As shown in Section~\ref{sec:privacy}, we need to estimate an expectation to measure the privacy, thus lower dimension will help us to avoid the curse of dimensionality during the approximation process.
	\item Reducing the dimension of the private-feature will intrinsically improve privacy as suggested by~\cite{osia2017_2} and~\cite{malekzadeh2017}.
\end{itemize}
Nevertheless, a potential disadvantage of the dimensionality reduction is that it can negatively affect on the accuracy of the primary variable prediction. However, we show in our experiments that the adverse effect of dimensionality reduction is negligible.

Reducing the dimensionality can be done as a preprocessing step on the pre-trained network. In fact, after choosing the intermediate layer, we can first execute the following operations: (i) Embed an auto-encoder with a low dimensional hidden layer on top of the chosen layer; (ii) Fine-tune the model to obtain the new primary variable predictor, and (iii) Choose the auto-encoder's hidden layer as the new intermediate layer which is low dimensional. Consequently, we can fine-tune the model with \emph{DPFE} architecture to get a low dimensional private-feature.

\subsection{Noise Addition} \label{sec:noise}

As mentioned earlier, many of the privacy preservation methods, from randomization technique to differentially private algorithms, rely on noise addition to gain privacy as it increases the uncertainty. We can utilize this technique after finishing the training procedure, in the test phase, when the dimensionality reduction is employed and the granularity of the sensitive variable is finer than the primary variable (e.g. identity is finer than gender). 

Adding noise to the private-feature will smooth out the conditional distributions of both primary and sensitive variables and form a tradeoff between privacy (of sensitive variable) and accuracy (of primary variable). This tradeoff can be helpful in real world applications, because one can choose the desired point on privacy-accuracy curve, based on the the importance of privacy or accuracy in a specific application. We will discuss this tradeoff in detail in Section~\ref{sec:eval}.

\section{Privacy Measure}\label{sec:privacy}


In this section, we propose a method for evaluating the quality of privacy algorithms. Considering the problem formulation by mutual information (Equation~\ref{eq:base}), one may suggest the negative of mutual information between the extracted private-feature and the sensitive variable ($-I(\textbf{f};\textbf{y})$), as a privacy measure. Since $I(\textbf{f};\textbf{y})=H(\textbf{y}) - H(\textbf{y}|\textbf{f})$ and $H(\textbf{y})$ is constant, this approach is equivalent to considering $H(\textbf{y}|\textbf{f})$ as the privacy measure. However, this measure has two shortcomings: (i) it is difficult to obtain an efficient estimation of $p(y|f)$; and (ii) there is no intuitive interpretation of this measure for privacy. In order to resolve these problems, we can relax the definition of uncertainty. We achieve this by partitioning the conditional probabilities by their rank order and build a lower bound for the conditional entropy:

\begin{align*}
\begin{split}
& H(\textbf{y}|\textbf{f}) = \int p(f) \sum_{y_a=1}^{c_y} p(y_a|f) \log \frac{1}{p(y_a|f)} df 
\end{split}
\end{align*}

It is known that among some numbers which sums into one, the $r$'th highest value is lower than or equal to $\frac{1}{r}$. So if we consider $r_{f,a}$ as the rank of $p(y_a|f)$ in the set of $\big\{p(y_j|f) | j \in \{1,\ldots,c_y\}\big\}$ sorted descending, we have:

\begin{align}
\begin{split}
 H(\textbf{y}|\textbf{f}) & \geq \int p(f) \sum_{y_a=1}^{c_y} p(y_a|f) \log r_{f,a} df \\
& = \mathds{E}_{p(f,y)} \log \textbf{r} \triangleq \mathcal{L}_{rank}
\end{split}
\end{align}
which leads to the following definition by dividing all formulas by $\log c_y$, in order to have a normalized measure between zero and one.

\begin{definition} [Log-Rank Privacy]
The \emph{log-rank privacy} of a discrete sensitive variable $\textbf{y}$, given the observed feature vector $\textbf{f}$, is defined as:
\begin{align}
LRP(\emph{\textbf{y}}|\emph{\textbf{f}}) = \frac{1}{\log c_y} \mathds{E}_{p(f,y)} \log \emph{\textbf{r}}
\end{align}
where $\emph{\textbf{r}}$ is a random function of $\emph{\textbf{f}}$, and $\emph{\textbf{y}}$ corresponds to the rank of $p(y|f)$ in the set of $\big\{p(y_j|f) | j \in \{1,\ldots,c_y\}\big\}$ which has been sorted in a descending order.
\end{definition} 

Assuming we have an estimation of $p(y|f)$, log-rank privacy can be empirically estimated by the sample mean of training in the rank logarithm:
\begin{align*}
\hat{LRP}(\textbf{y}|\textbf{f}) &= \frac{1}{N \log c_y}\; \sum_{i=1}^N \ \log\Big(rank\big(p(y_i|f_i), S_i\big)\Big) \\
S_i &= \big\{p(y_j|f_i) | j \in \{1,\ldots,c_y\}\big\}
\end{align*}
where $rank(a,S)$ for $a \in S$ is the rank of $a$ in the descending ordered set $S$. In the following, we provide some intuition about the log-rank privacy and its relation to entropy, k-anonymity and classification error.\\

\textbf{20-questions game interpretation.} Consider the 20-questions game, in which we want to guess an unknown object by asking yes/no questions from an oracle. As stated in~\cite{cover2012}, the entropy is equivalent to the minimum number of questions one could ask in order to find the correct answer. Now consider the situation where we cannot ask \emph{any} kind of yes/no questions, but only the questions in which we guess to be a candidate for the final answer, e.g. 'is the answer a chair?'. Also assume that if we can guess the correct answer after $k$ questions, we would be penalized by $\log k$; so that the wrong guesses are punished more at the beginning. Evidently, the optimal strategy to have the minimum expected penalty is to guess the objects in the same order with their probabilities. Using this strategy, the expected penalty would be equal to the \emph{log-rank privacy}. \\


\textbf{k-anonymity and expected rank.} k-anonymity deals with the number of entities we are equally uncertain about. Expected rank can be considered as a soft interpretation of this number, relaxing the equal uncertainty with the weighted sum of ranks. Thus, the rank variable expectation can be thought as the expected number of entities that we are in doubt about. \\

\textbf{Classification error extension.} One could suggest using \emph{classification error (zero-one loss)} as the privacy measure, as it represents the deficiency of the classifier. Using this measure is equal to considering zero and one penalty for the correct and wrong guesses of the \emph{first} question, respectively. Thus, two situations where we can find the correct label in the second and tenth question are considered equal and both penalized by one. The log-rank privacy handles this issue by penalizing different questions using their ranks' logarithm and can be considered as an extension of classification error. \\

\textbf{Sensitivity analysis.} Empirically approximating an expected value by drawing samples from a probability distribution is a common method in machine learning \cite{bishop2006}. For comparing empirical estimation of log-rank privacy with entropy, we need to estimate the order of probabilities in the former, while the exact values of probabilities are needed in the later. In general, approximating the log-rank privacy is less sensitive to the error of the density estimation and can gain lower variance. Detailed sensitivity analysis is out of scope of this paper and will be considered in future work. 
\section{Evaluation}
\label{sec:eval}


In this section, we evaluate the proposed private-feature extractor by considering the problem of \emph{facial attribute} prediction. We use each face image as an input and infer its facial attributes such as gender, expression, or age, in a supervised manner. We extract a feature for facial attribute prediction, which at the same time is non-informative with respect to the \emph{identity} of the person (sensitive attribute). In all of our experiments, we used the CelebA face dataset, presented in~\cite{liu2015}, which includes 40 binary facial attributes, such as gender (male/female), age (young/old), and smiling (yes/no) with the corresponding identity labels. In the following, first we explain the experiment setting and then we discuss the results.

\subsection{Experiment Setting}

\begin{algorithm}[t] 
	\caption{DPFE Training Phase}
	\label{alg:train}
	\begin{algorithmic}
		\REQUIRE training data, intermediate layer, attribute set 
		\STATE $M_0 \leftarrow$ attribute prediction model
		\STATE $L \leftarrow \text{intermediate layer of } M_0$ (e.g. \emph{conv7})
		\STATE $|L| \leftarrow$ size of $L$'s output
		\STATE $A \leftarrow \text{attribute set}$ (e.g. \emph{\{Gender \& Age\}})
		\STATE $AE \leftarrow$ linear auto-encoder with input/output size of $|L|$
		\STATE $H \leftarrow$ hidden layer of $AE$ (private-feature layer)
		\STATE Initialize $AE$ with PCA weights on $L$'s output 
		\STATE $M_1 \leftarrow$ embed $AE$ to $M_0$ on top of $L$
		\STATE $S_{L,A} \leftarrow$ fine-tune $M_1$ on $A$
		\STATE $z \leftarrow A\;,\; y \leftarrow Identity$
		\STATE $P_{L,A} \leftarrow$ fine-tune $S_{L,A}$ with \emph{DPFE} architecture
		\ENSURE $S_{L,A}$: simple model, $P_{L,A}$: \emph{DPFE} fine-tuned model, $H$: private-feature layer
	\end{algorithmic}
\end{algorithm}

In our evaluations, we used the layer separation mechanism followed by dimensionality reduction and noise addition. We selected the state of the art pre-trained facial attribute prediction model presented in~\cite{rudd2016} and called it the \emph{original model}.\footnote{We used a similar implementation from \url{https://github.com/camel007/caffe-moon} which used the tiny darknet architecture from \url{https://pjreddie.com/darknet/tiny-darknet/}.} Then, we chose an attribute set (e.g. \{gender \& age\}) to preserve its information as the private-feature. Next, we selected an intermediate layer (e.g. layer conv7) of the chosen network. Since this layer can also be a high dimensional tensor, we embeded a linear auto-encoder and applied batch normalization on its hidden layer to obtain the normalized intermediate features. Finally, by fine-tuning the network, we may have an attribute prediction model with low dimensional intermediate feature, which we refer to as the \emph{Simple model} in the rest of the paper.
While the low-dimensional feature preserves the information of attributes (see Theorem~\ref{th0}), it does not necessarily omit the sensitive information. Hence, we should fine-tune the network with the proposed \emph{DPFE} architecture (figure~\ref{fig:arc}) to remove identity information from the intermediate features. We refer to this model as the \emph{DPFE model}. These steps are depicted in Procedure~\ref{alg:train}. We implemented all the models with the \emph{Caffe} framework \cite{jia2014caffe}, utilizing the \emph{Adam} optimizer \cite{kingma2014}, and a contrastive loss function.

We evaluated each fine-tuned model based on the following criteria:
\begin{itemize}
\item Accuracy of the facial attribute prediction: achieving higher accuracy implies that the primary variable information is well preserved.
\item Identity privacy: we evaluate the privacy of the feature extractor, using two different measures. First, the log-rank privacy measure, introduced in Section~\ref{sec:privacy}. Second, we utilize 1NN identity classifier and consider its misclassification rate which must be high in order to preserve privacy (although this condition is not sufficient as discussed in Section~\ref{sec:privacy}. We also use the deep visualization technique presented in~\cite{dosovitskiy2016} to demonstrate that the higher layers of the deep network may not be reliable.
\end{itemize}



To show the generality of the proposed method, we consider four different intermediate layers (conv4-2, conv5-1, conv6-1 and conv7) together with five attribute sets (listed below), and the results for twenty \emph{Simple} and twenty \emph{DPFE} models. 
\begin{itemize}
	\item G: \{gender\}
	\item GA: \{gender, age\}
	\item GAS: \{gender, age, smiling\}
	\item GASL: \{gender, age, smiling, big~lips\}
	\item GASLN: \{gender, age, smiling, big~lips, big~nose\}
\end{itemize}

In what follows, we first explain the accuracy-privacy tradeoff based on the log-rank privacy measure and 1NN misclassification rate (Subsection~\ref{sec:accpriv}). We then present the visualization result (Subsection~\ref{sec:vis}), and finally address the complexity issue of the private-feature extractor by implementing the proposed framework on a smartphone (Subsection~\ref{sec:mobile}).

\subsection{Accuracy \emph{vs.} Privacy} \label{sec:accpriv}

To evaluate \emph{Simple} and \emph{DPFE} models, we designed the following four experiments and assessed different models based on their accuracy-privacy trade-off:

\begin{enumerate}
	\item We compared \emph{Simple} and \emph{DPFE} models to show the superiority of \emph{DPFE} fine-tuning;
	\item We assessed the effect of different intermediate layers to indicate the appropriateness of higher layers;
	\item We evaluated the effect of extending attribute set and showed that preserving privacy becomes harder; 
	\item We considered mean and standard deviation of Rank-privacy measure to guarantee privacy.
	
\end{enumerate}


\begin{algorithm}[t]
	\caption{DPFE Test Phase}
	\label{alg:test}
	\begin{algorithmic}
		\REQUIRE test data, intermediate and private-feature layers, attribute set, model
		\STATE $H \leftarrow$ private-feature layer
		\STATE $A \leftarrow$ attribute set
		\STATE $M \leftarrow$  model
		\STATE $C \leftarrow$ covariance matrix of $H$ in $M$
		\FOR{$r \in \{ratios\}$}
		\STATE $N_r \leftarrow$ Gaussian noise layer with covariance $rC$
		\STATE $M_r \leftarrow$ embed $N_r$ as an additive noise on $H$ in $M$
		\STATE $H_r \leftarrow$ output of $H + N_r$
		\STATE $p_r \leftarrow$ \emph{identity} privacy of $H_r$ 
		\STATE $a_r \leftarrow$ average accuracy of $M_r$ on $A$
		\ENDFOR
		\STATE plot accuracy-privacy curve using $\big\{(a_r,p_r)|r \in \{ratios\}\big\}$
		\ENSURE accuracy-privacy trade-off
	\end{algorithmic}
\end{algorithm}

In order to adjust the accuracy-privacy trade-off, we used the noise addition mechanism. After the training phase, we estimate the covariance matrix of the feature space, scale it with different ratios and use it as a covariance matrix of a Gaussian noise. By increasing the amount of noise, the accuracy of the primary variable prediction decreases but the privacy of the sensitive variable increases. As a result, we can build the accuracy-privacy trade-off curves in a manner similar to the trade-off in rate-distortion theory (see~\cite{cover2012}). The evaluation steps are shown in Procedure~\ref{alg:test}. The accuracy-privacy curves of different models can be compared based on the following definition.

\begin{definition} [Acc-Priv superiority] For two models that try to preserve privacy of a sensitive variable and maintain accuracy of a primary variable, the one which always results in higher value of privacy for a fixed value of accuracy, is \emph{Acc-Priv superior}.
\end{definition}

Considering Equation~\ref{eq:relaxed}, it seems that the relative importance of accuracy and privacy can be controlled by changing the values of parameter $\beta$. However, this is not feasible in practice due to the challenges in the training stage. For example, training with a constant $\beta$ and consequent noise addition mechanism, it is possible to set different accuracy-privacy strategies by utilizing a single trained model. This is not the case when we have various models by considering different values for $\beta$. We used cross validation, in order to choose a suitable fixed value for $\beta$ in our experiments. 

We computed the accuracy-privacy trade-off on the test data with 608 identities. Setting noise to zero, for all intermediate layers and attribute sets, \emph{Simple} and \emph{DPFE} models reached the same accuracy level as the \emph{original} model with an error margin of less than $0.5\%$.\footnote{In order to report the accuracy of an attribute set, we consider the average accuracy of predicting each binary attributes in the set.} Therefore, we can conclude that all \emph{Simple} and \emph{DPFE} models preserve the facial attribute information, and we may concentrate on their privacy performance. \\

\begin{figure*}[!ht] 
	\centering
	\begin{subfigure}[t]{0.25\textwidth}
		\centering
		\includegraphics[width=\linewidth]{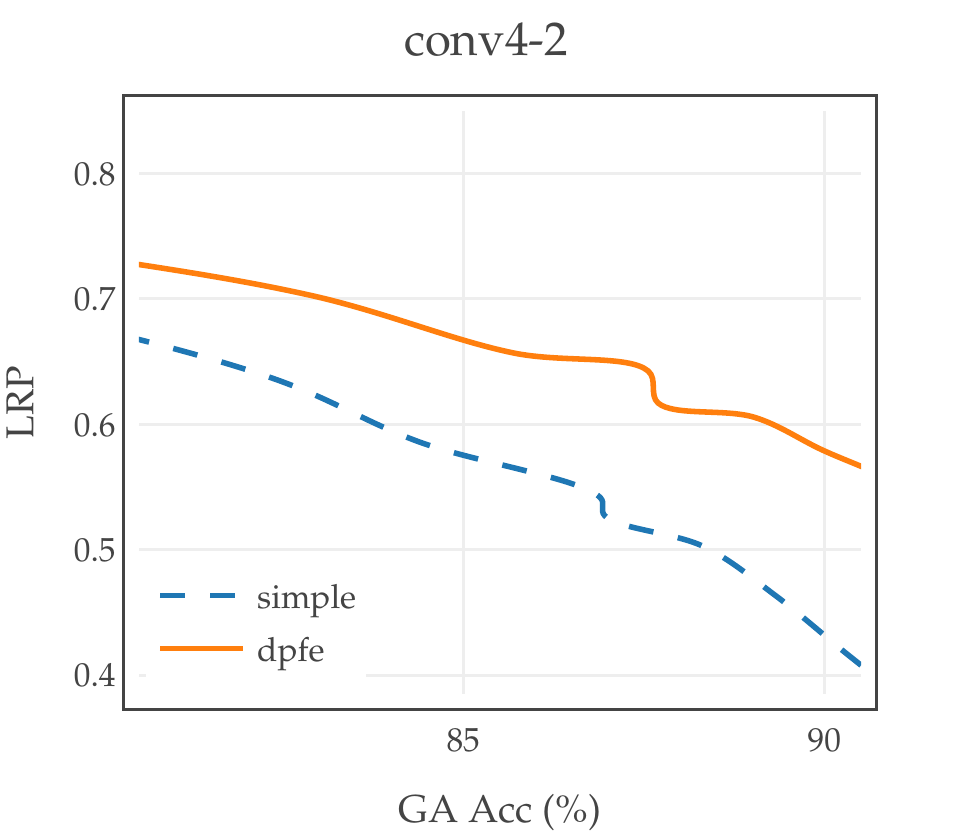}
	\end{subfigure}%
	\begin{subfigure}[t]{0.25\textwidth}
		\centering
		\includegraphics[width=\linewidth]{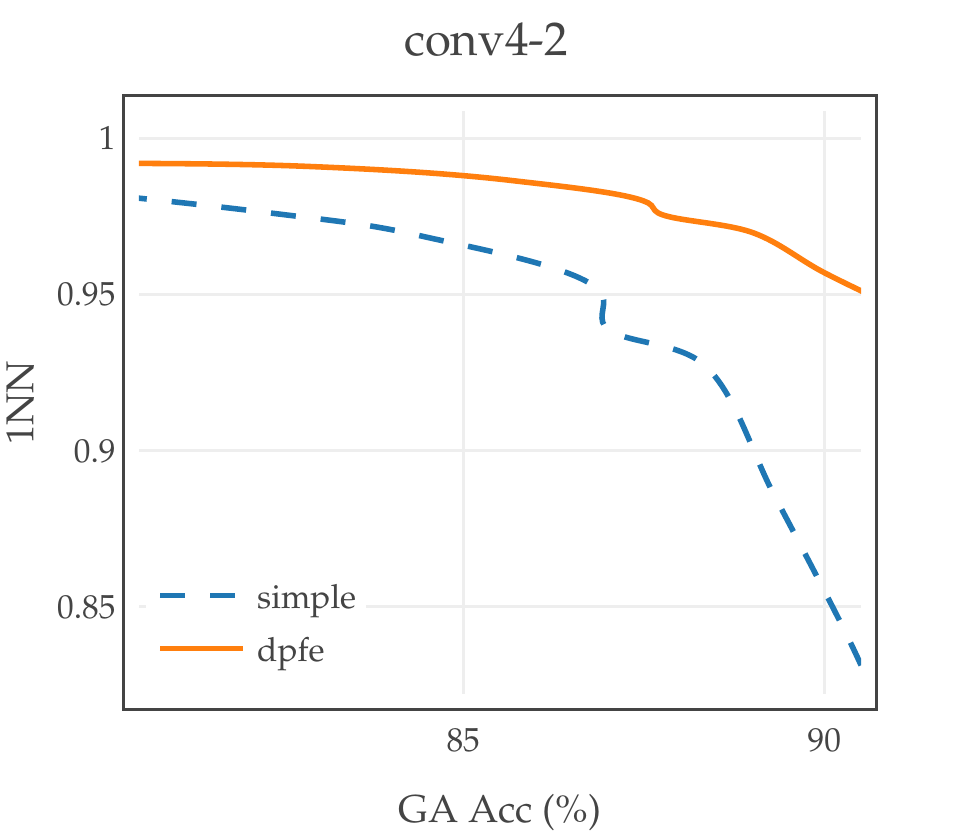}
	\end{subfigure}%
	\begin{subfigure}[t]{0.25\textwidth}
		\centering
		\includegraphics[width=\linewidth]{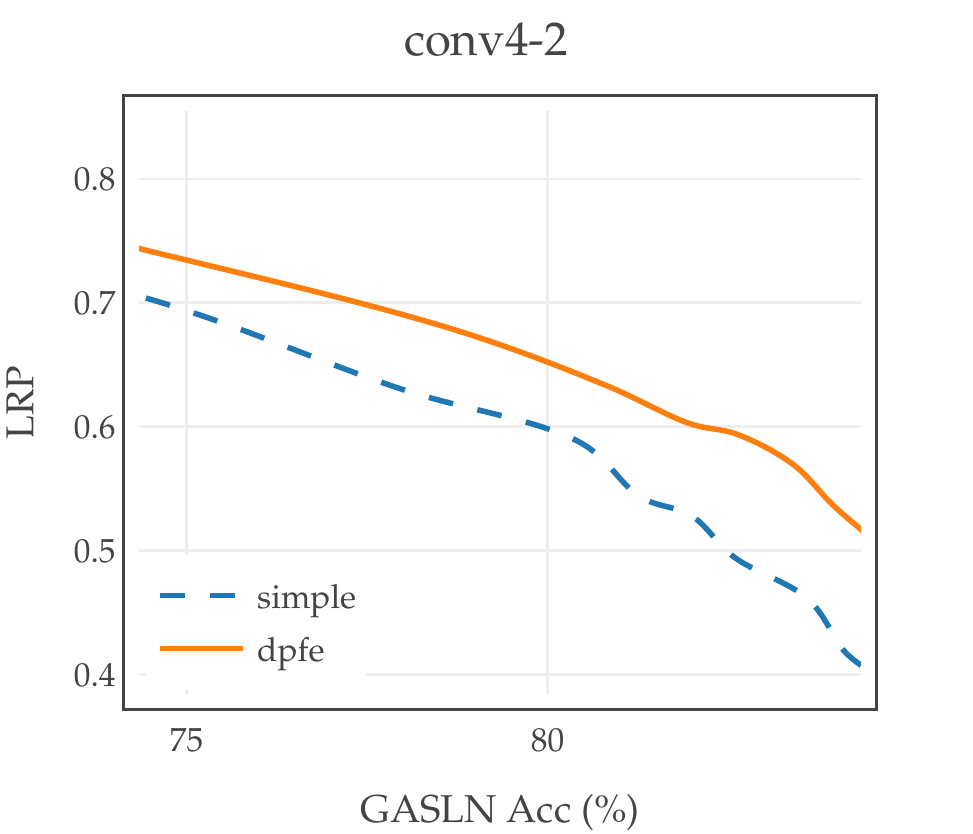}
	\end{subfigure}%
	\begin{subfigure}[t]{0.25\textwidth}
		\centering
		\includegraphics[width=\linewidth]{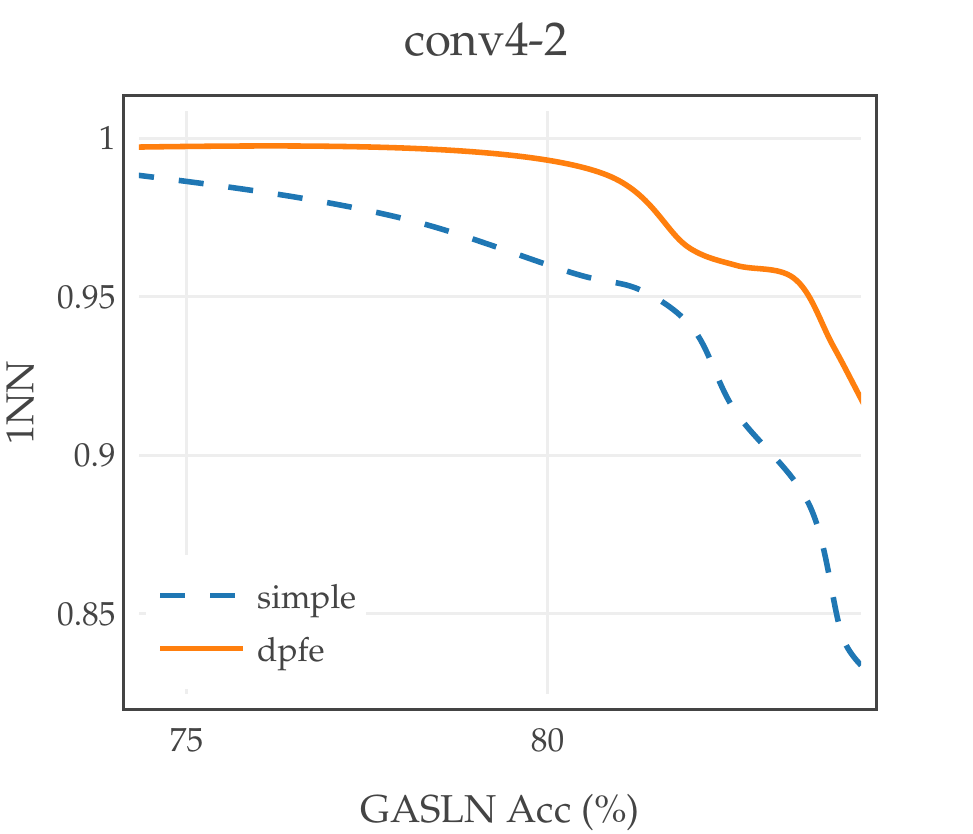}
	\end{subfigure}
	\\
	\vspace{15pt}
	\begin{subfigure}[t]{0.25\textwidth}
		\centering
		\includegraphics[width=\linewidth]{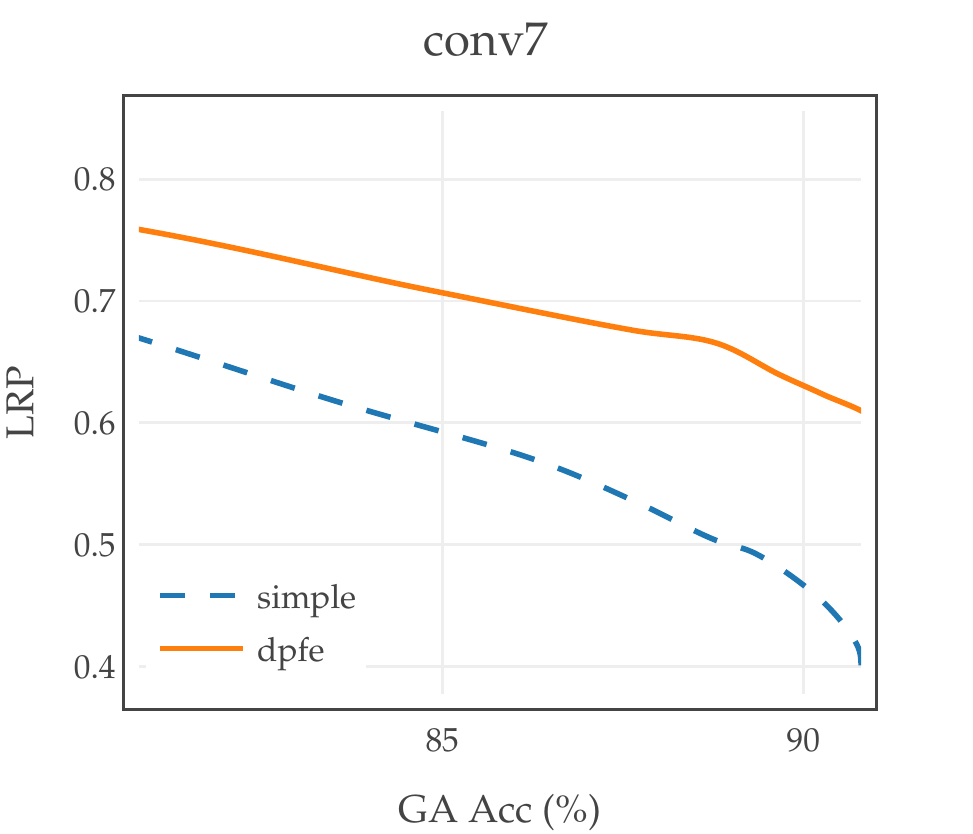}
	\end{subfigure}%
	\begin{subfigure}[t]{0.25\textwidth}
		\centering
		\includegraphics[width=\linewidth]{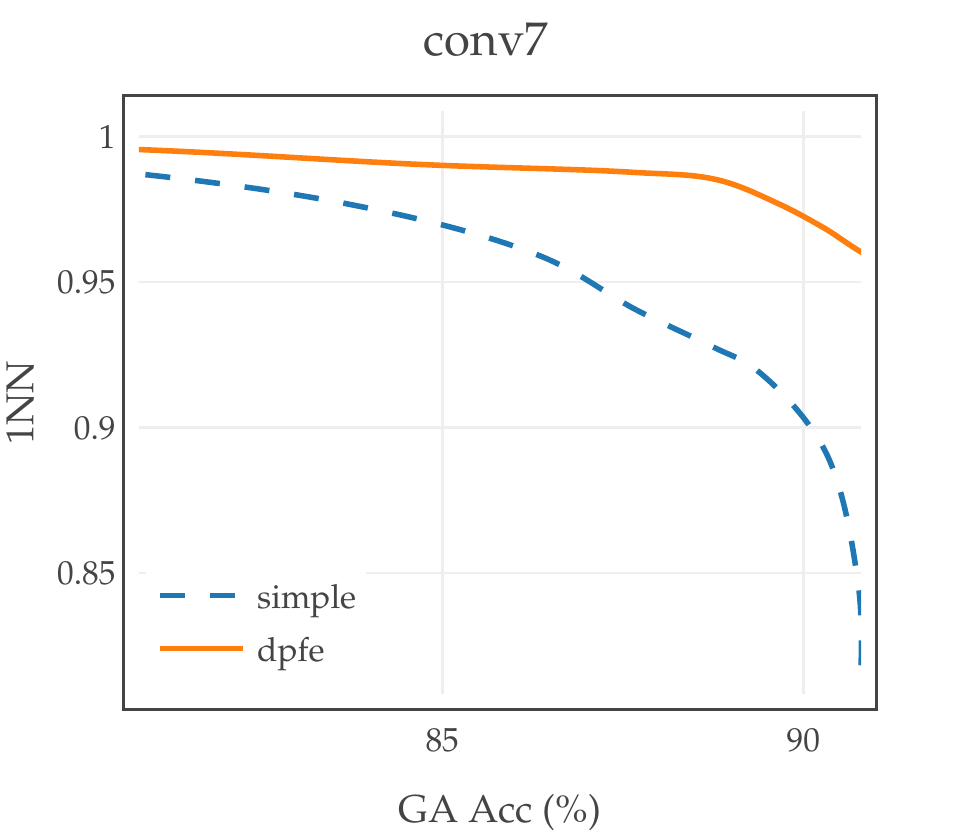}
	\end{subfigure}%
	\begin{subfigure}[t]{0.25\textwidth}
		\centering
		\includegraphics[width=\linewidth]{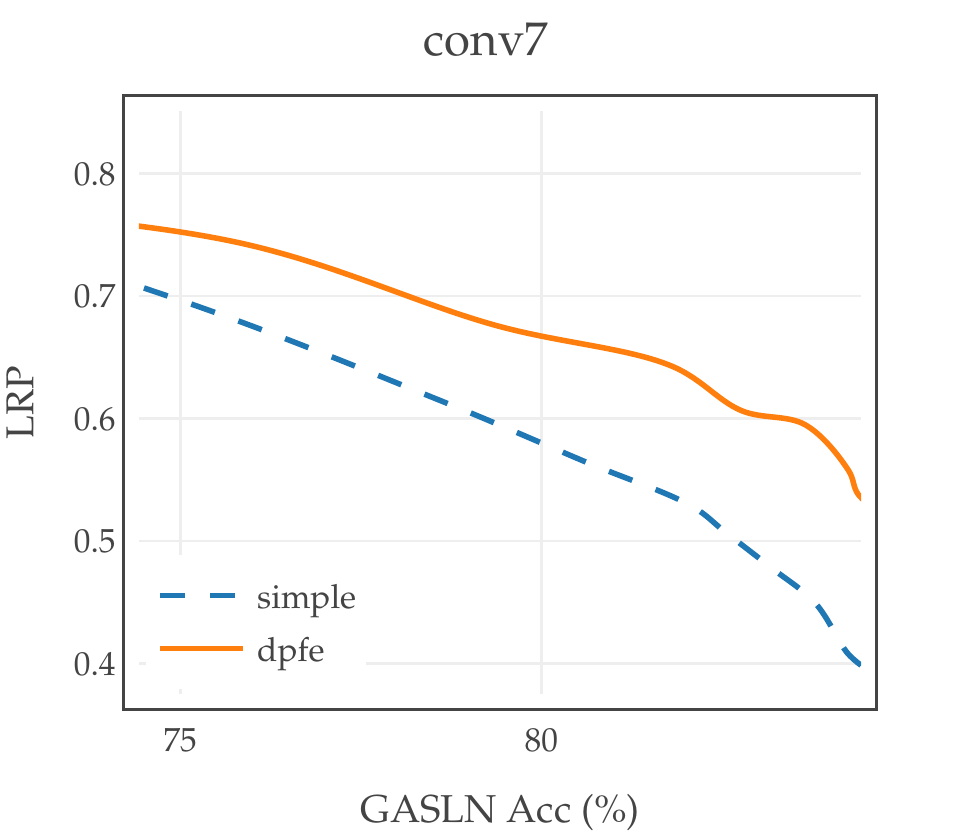}
	\end{subfigure}%
	\begin{subfigure}[t]{0.25\textwidth}
		\centering
		\includegraphics[width=\linewidth]{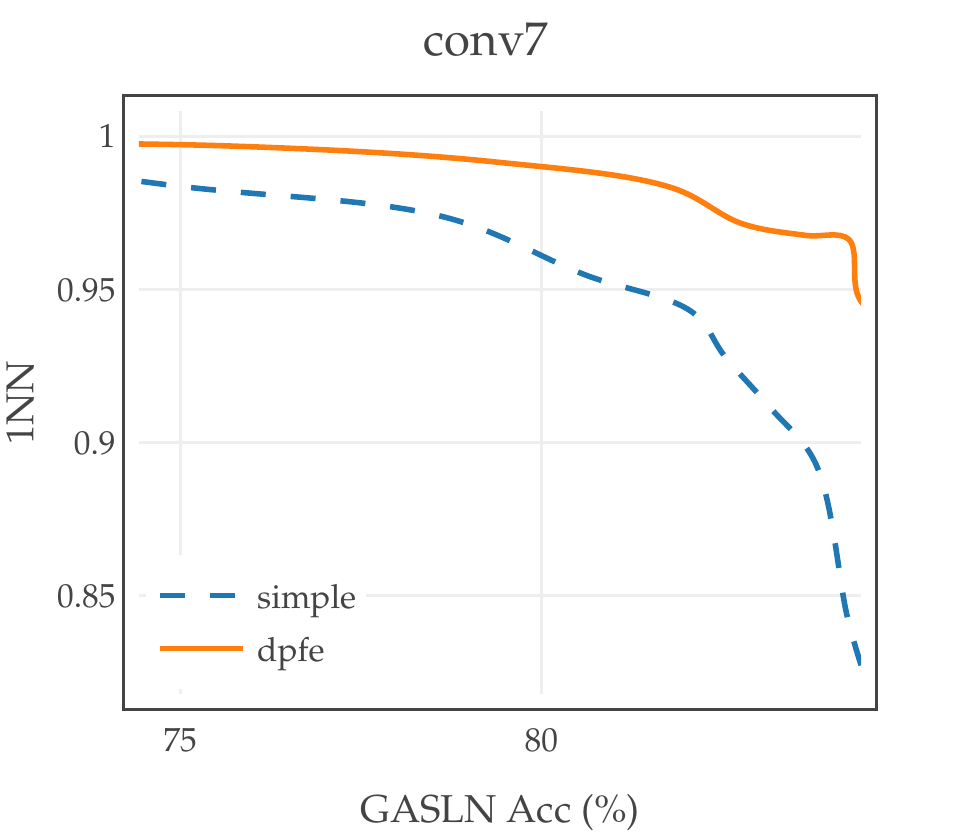}
	\end{subfigure}
	\caption{\emph{DPFE} \emph{vs.} Simple models: fine-tuned models with \emph{DPFE} architecture achieve Acc-Priv superiority to corresponding Simple models in all layers and attribute sets.}
	\label{fig:dpfe_vs_simple}
\end{figure*}

\textbf{Effect of \emph{DPFE fine-tuning}.} In order to verify the superiority of \emph{DPFE} fine-tuning over \emph{Simple} fine-tuning, we compared the accuracy-privacy curve of different models, fine-tuned with DPFE or Simple architectures. Figure~\ref{fig:dpfe_vs_simple} shows the results for the combination of two layers and two attribute sets, with different privacy measures. In all cases, DPFE models have the Acc-Priv superiority over Simple models. In other words, for a fixed value of accuracy, DPFE consistently achieves higher levels of privacy.\\

\begin{figure*}[!ht] 
	\centering
	\begin{subfigure}[t]{0.25\textwidth}
		\centering
		\includegraphics[width=\linewidth]{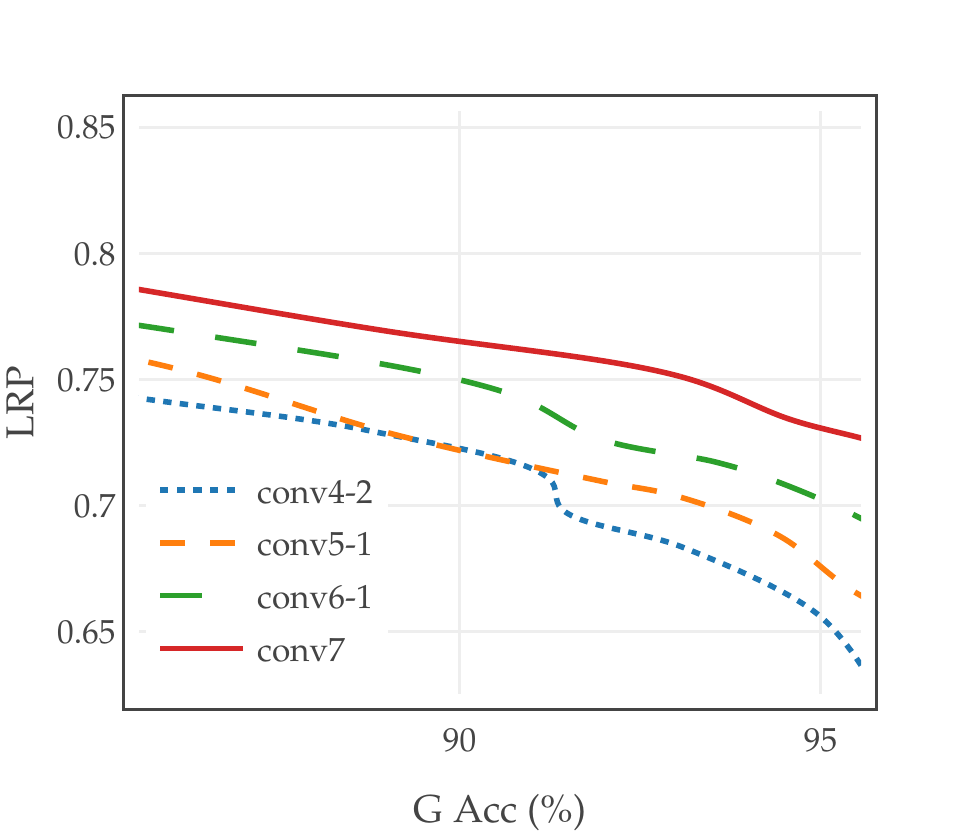}
	\end{subfigure}%
	\begin{subfigure}[t]{0.25\textwidth}
		\centering
		\includegraphics[width=\linewidth]{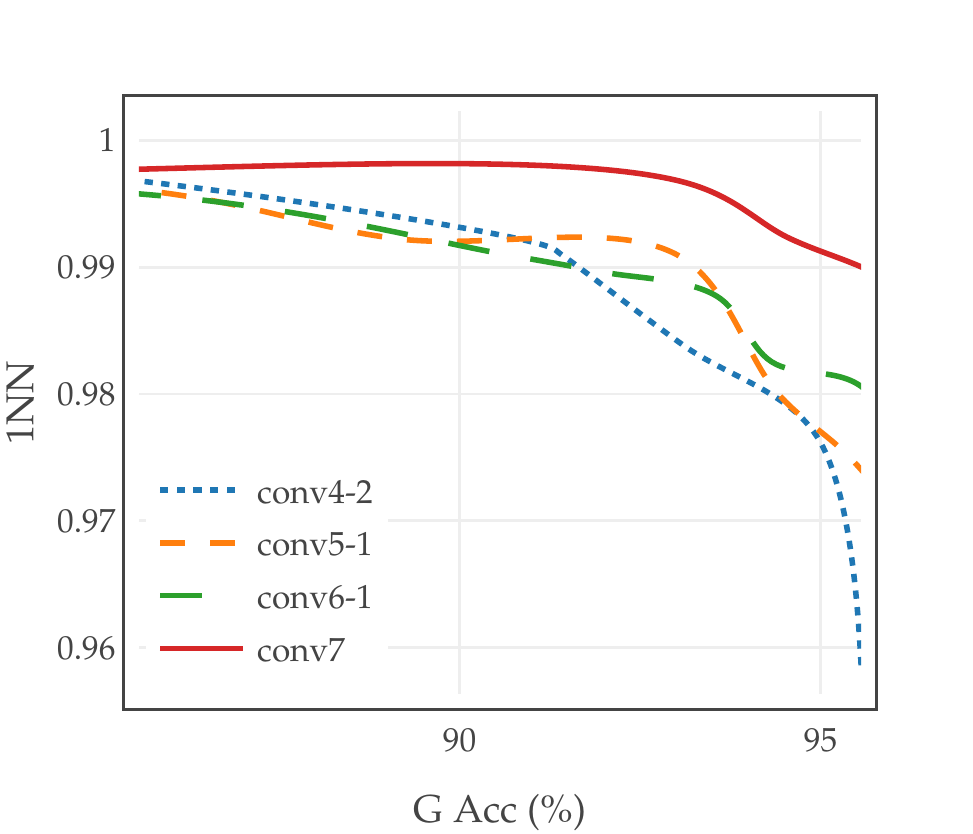}
	\end{subfigure}%
	\begin{subfigure}[t]{0.25\textwidth}
		\centering
		\includegraphics[width=\linewidth]{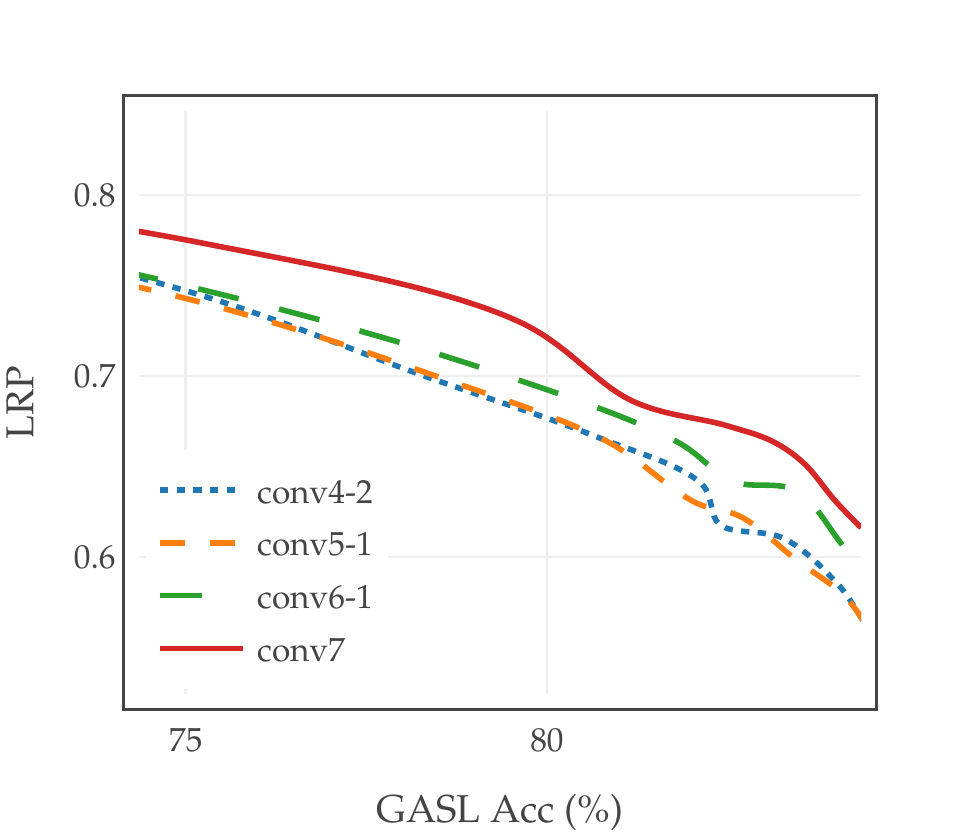}
	\end{subfigure}%
	\begin{subfigure}[t]{0.25\textwidth}
		\centering
		\includegraphics[width=\linewidth]{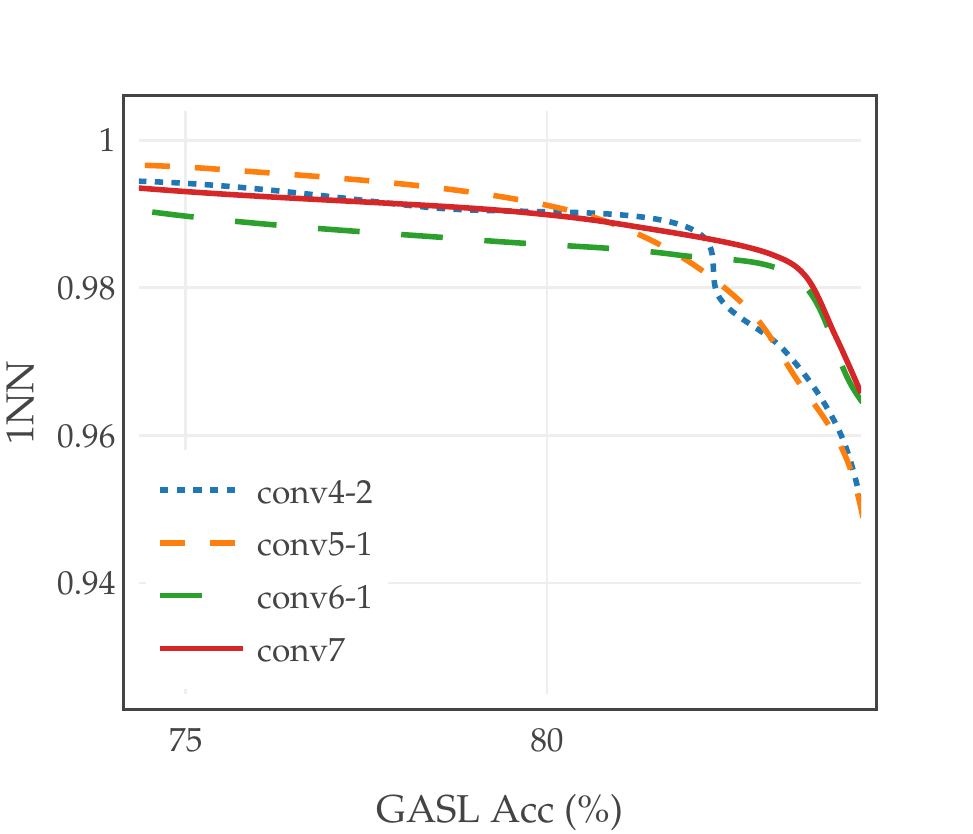}
	\end{subfigure}
	\caption{Layer Comparison: in general, higher layers achieve Acc-Priv superiority to lower layers. In this figure, all models are fine-tuned with the \emph{DPFE} architecture.}
	\label{fig:layer_comparison}
\end{figure*}

\textbf{Effect of higher layers.} Comparison of the accuracy-privacy curves of different layers on the same attribute set is depicted in Figure~\ref{fig:layer_comparison}. The results illustrate the Acc-Priv superiority of higher layers for two attribute sets and for both privacy measures. This observation is inline with our earlier assumptions about the higher layers.\\

\begin{figure}[t]
	\centering
	\begin{subfigure}[t]{0.5\columnwidth}
		\centering
		\includegraphics[width=\linewidth]{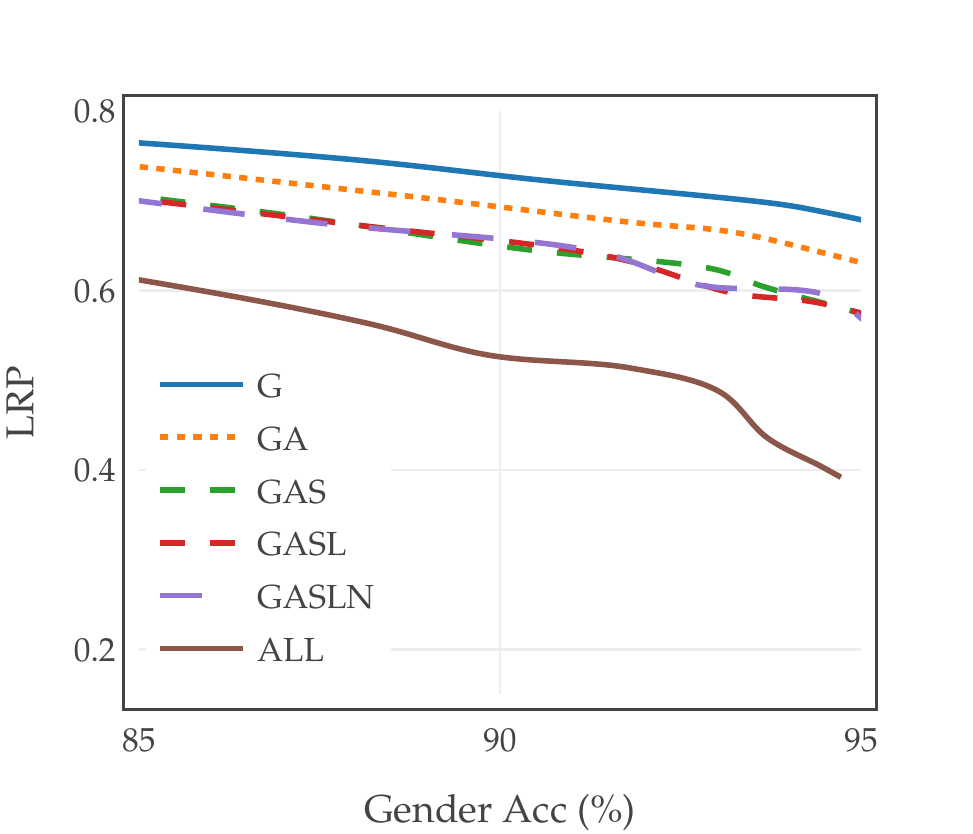}
	\end{subfigure}%
	\begin{subfigure}[t]{0.5\columnwidth}
		\centering
		\includegraphics[width=\linewidth]{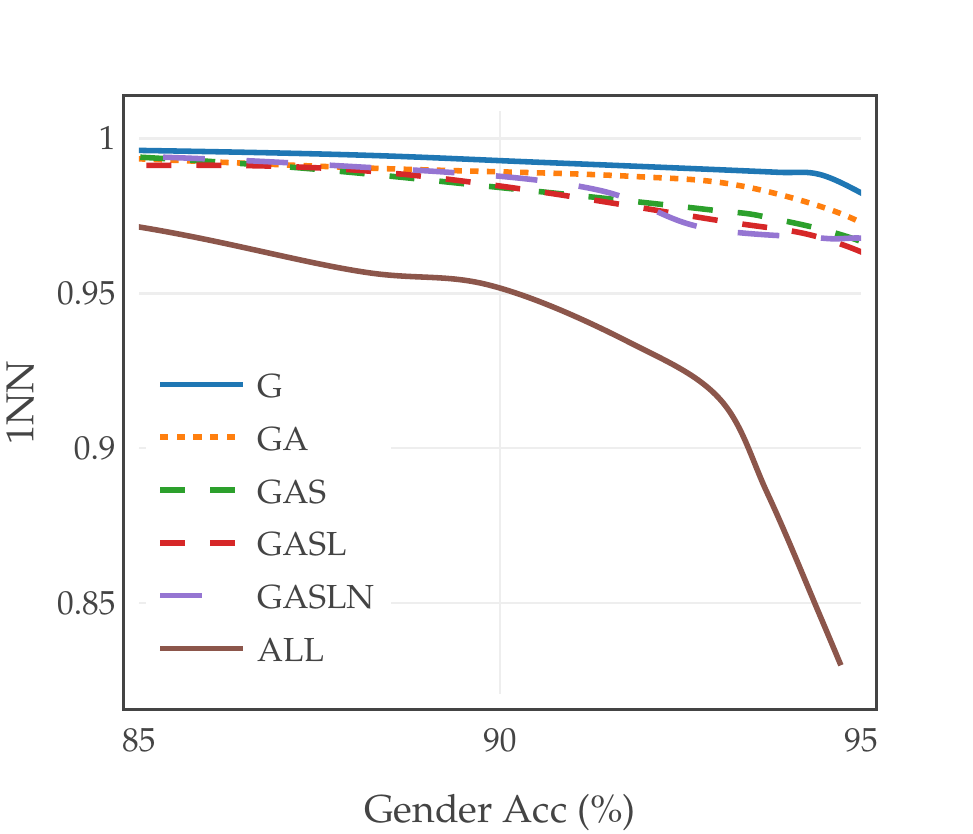}
	\end{subfigure}
	\caption{Comparison of Gender accuracy-privacy trade-offs when putting more preservation constraints on the model. The intermediate layer is set to conv7. }
	\label{fig:specificity}
\end{figure}

\textbf{Effect of attribute set extension.} The accuracy-privacy trade-off of the \emph{DPFE} fine-tuned models for different attribute sets with conv7 as the intermediate layer, are shown in figure~\ref{fig:specificity}. The results show that as we enlarge the attribute set and restrict the model with preserving the information, then preserving privacy becomes more challenging due to the intrinsic correlation of the identity with facial attributes.\\ 

\begin{figure}[!h]
	\centering
	\includegraphics[width=\columnwidth]{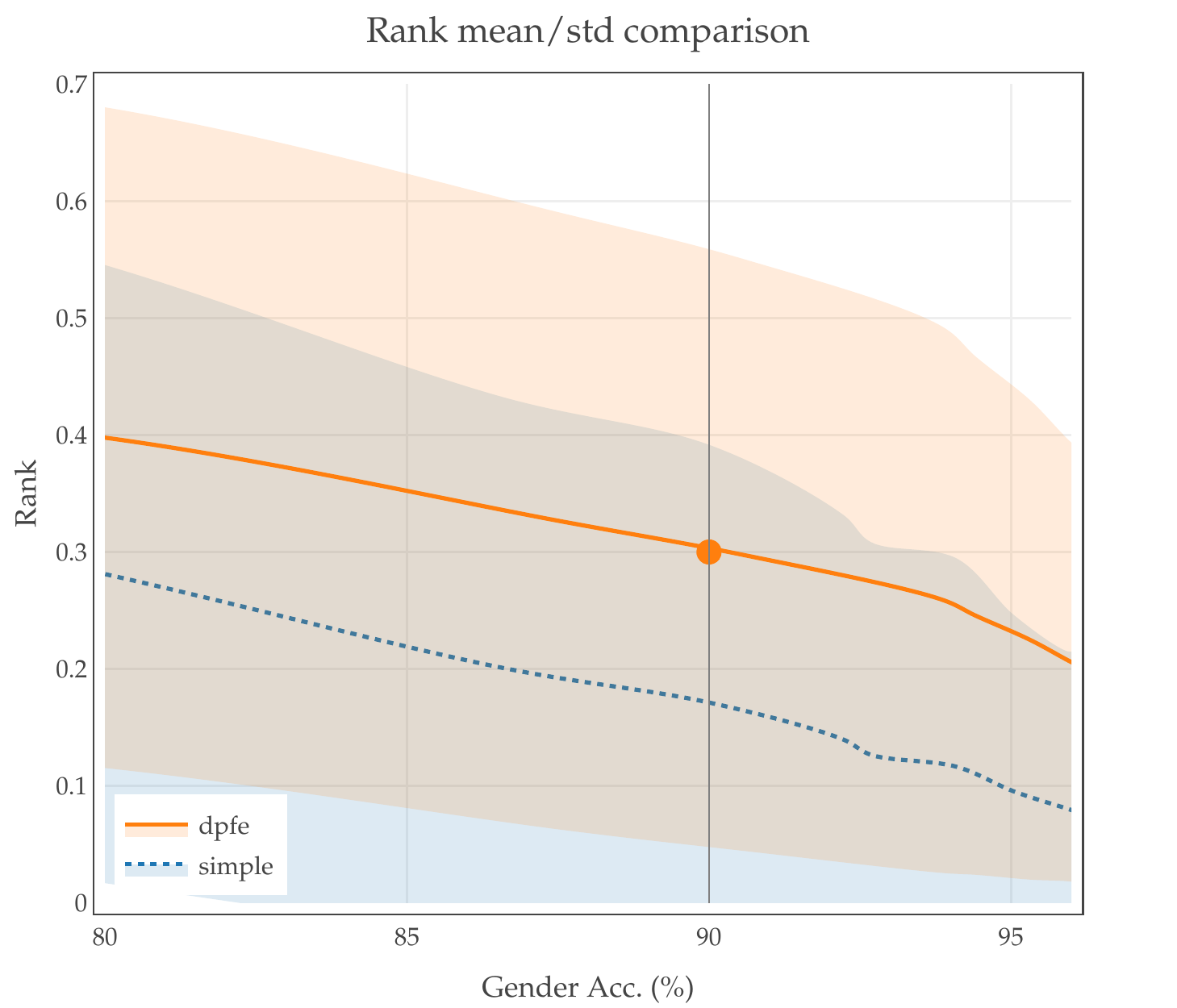}
	\caption{Comparison of mean and standard deviation of Rank variable for \emph{DPFE} and Simple models for layer conv7.}
	\label{fig:std_comparison}
\end{figure}

\textbf{Guaranteeing privacy.} As discussed in Section~\ref{sec:privacy}, instead of log-rank, we could also consider the rank itself by analyzing its mean and variance. This idea is depicted in figure~\ref{fig:std_comparison} for Simple and \emph{DPFE} models. The results show that the \emph{DPFE} model has Acc-Priv superiority over the Simple model. More importantly, it forces the conditional distribution of the sensitive variable to converge to an uniform distribution, at least in the rank-mean and standard deviation sense. In fact, the mean and the standard deviation of rank measure for the discrete uniform distribution are $0.5$ and $0.28$, respectively. As shown in figure~\ref{fig:std_comparison}, when privacy increased, the statistics for the \emph{DPFE} model converge to their corresponding values for the uniform distribution. If we consider the normal distribution for the rank variable, we can provide an $(\epsilon,\delta)$ privacy guarantee, similar to the method used in differential privacy \cite{dwork06}. For example, as depicted in figure~\ref{fig:std_comparison}, we can achieve the gender accuracy of up to $90\%$ with a rank-mean of $0.3$ and standard deviation of $0.25$. Hence, with a probability of $0.88\%$ we can claim that the rank-privacy is greater than $0.1$, and we have achieved $10\%$ anonymity.


\subsection{Visualization} \label{sec:vis}

\begin{figure*}[ht]
	\centering
	\includegraphics[width=.9\linewidth]{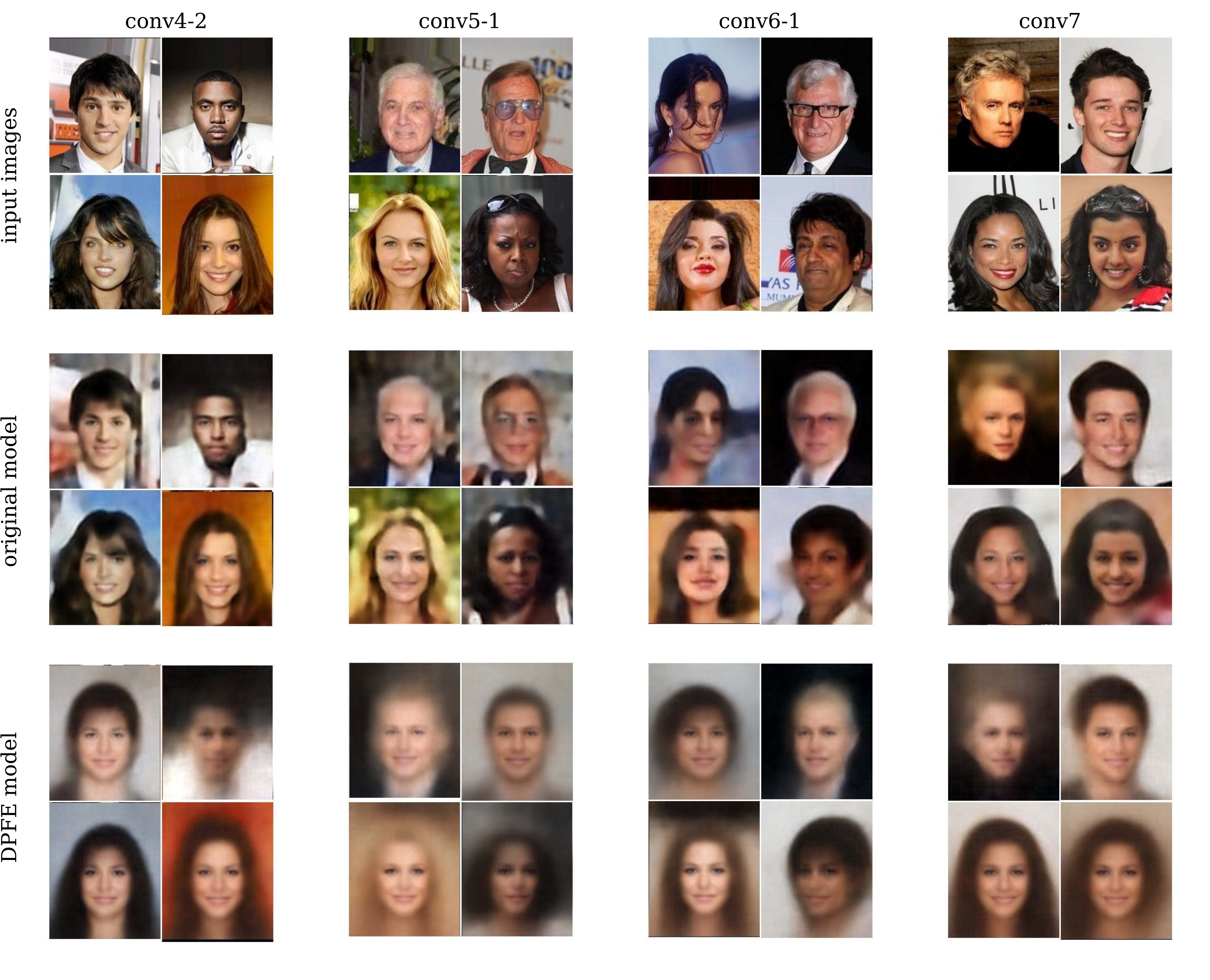}
	\caption{Visualization of different layers for different models: from top to bottom rows show input images, reconstructed images from original model and reconstructed images from \emph{DPFE} model. The second row shows that separating layers of a deep model and relying on specificity of higher layers does not provide identity privacy.}
	\label{fig:vis}
\end{figure*}

Visualization is a method for understanding the behavior of deep networks. It provides an insightful intuition about the flow of information through different layers. We used an auto-encoder objective visualization technique \cite{dosovitskiy2016} to validate the sensitive information removal in DPFE. 
The reconstruction of images is done by feeding the private-feature to the Alexnet decoder proposed in~\cite{dosovitskiy2016}. Therefore, we may \emph{visually} verify the identity removal property of the private-feature by comparing the original and reconstructed images. These images are shown in figure~\ref{fig:vis} for different layers of the original and \emph{DPFE} fine-tuned models. 


The results can be analyzed in two aspects: accuracy of desired attributes and privacy of identities. From the privacy perspective, the identity of the people in the reconstructed images of the original model can be readily observed in the last layers (e.g. conv7), while that is not the case for \emph{DPFE} models. Therefore, just relying on the output of higher layers in the original model can not assure acceptable privacy preservation performance, while the \emph{DPFE} models assure the privacy of identities. Regarding the accuracy, we can observe and detect the facial attributes in both models. 


\subsection{Complexity \emph{vs.} Efficiency} \label{sec:mobile}


\begin{figure*}[t] 
	\centering
	\begin{subfigure}[t]{0.5\textwidth}
		\centering
		\includegraphics[width=\linewidth]{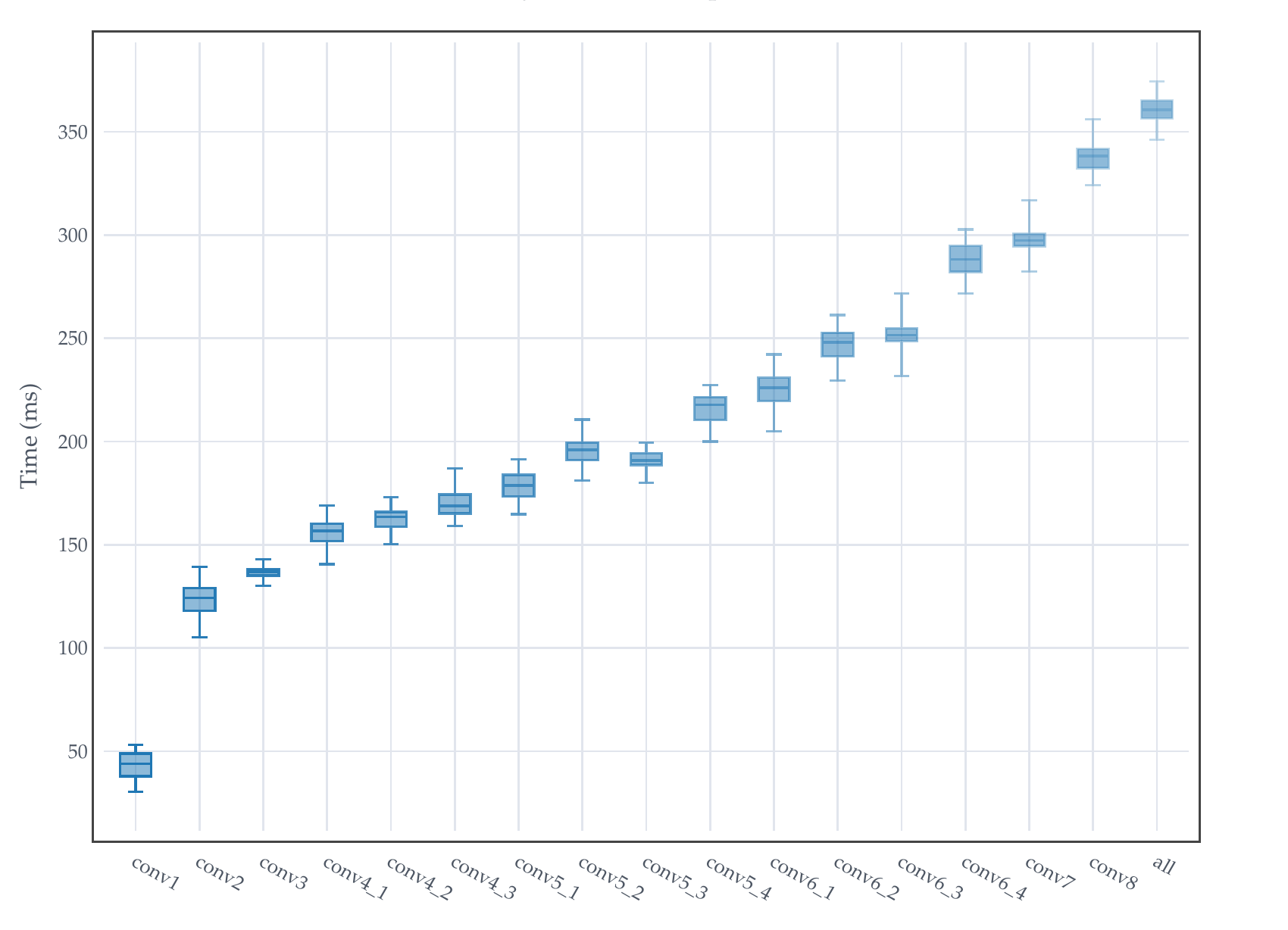}
		\caption{Layers time comparison}
		\label{fig:time_boxplot}
	\end{subfigure}%
	\begin{subfigure}[t]{0.5\textwidth}
		\centering
		\includegraphics[width=\linewidth]{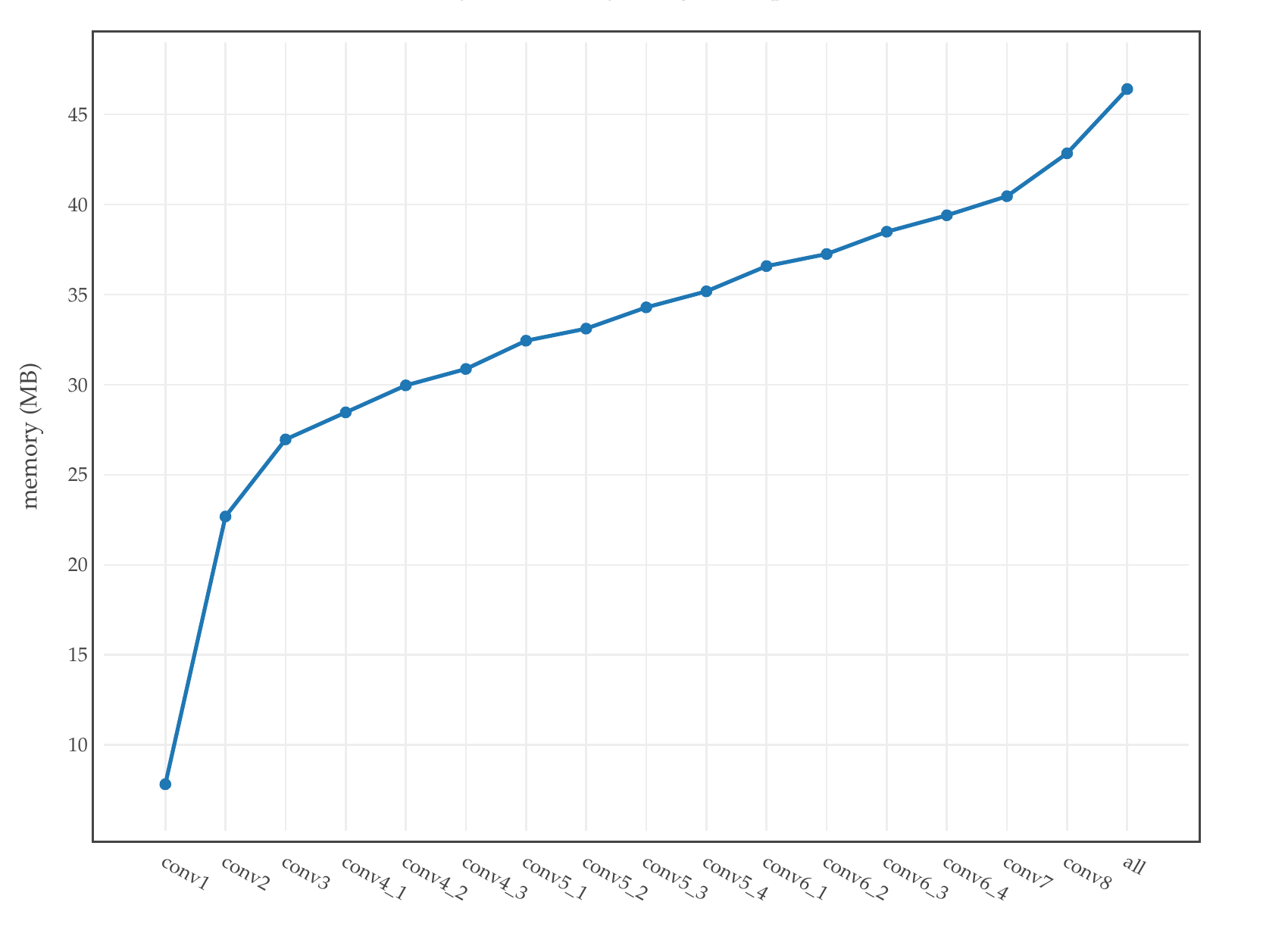}
		\caption{Layers memory usage comparison}
		\label{fig:mobile_memory}
	\end{subfigure}
	\caption{Comparison of different layers on mobile phone.}
	\label{fig:mobile}
\end{figure*}

\begin{table}[!b]
	\centering
	\caption{Device Specification}
	\label{tab:device_spec}
	\begin{tabular}{| l | l |}
		\hline
		\multicolumn{2}{|c|}{Google (Huawei) Nexus 6P} \\
		\hline
		Memory        & 3 GB LPDDR4 RAM \\
		Storage       & 32 GB \\
		CPU           & Octa-core Snapdragon 810 v2.1  \\
		GPU           & Adreno 430 \\
		OS            & Android 7.1.2 \\
		\hline
	\end{tabular}
\end{table}

Although higher intermediate layers may achieve better accuracy-privacy trade-off, in some cases, such as low-power IoT devices or smartphones, their computational complexity may not be acceptable. Therefore, due to the limited resources on these devices (both  memory and computational power) a privacy-complexity trade-off should also be considered. In order to address this problem, we evaluated the original architecture without dimensionality reduction on a smartphone and measured its complexity in different layers. The results are shown in figure~\ref{fig:mobile}. By gradually reducing the complexity of the private-feature extractor (considering lower intermediate layers in the layer separation mechanism), we also managed to reduce the inference time, memory and CPU usage, while hiding the user's sensitive information.

We evaluated the proposed implementation on a modern handset device, as shown in Table~\ref{tab:device_spec}. We evaluated the intermediate layers cumulatively, and compared them with the on-premise solution (full model). We used Caffe Mobile~v1.0~\cite{jia2014caffe} for Android to load each model and measured the inference time (figure~\ref{fig:time_boxplot}) and model memory usage (figure~\ref{fig:mobile_memory}) of each of the 17 configurations. We configured the model to only use one core of the device's CPU, as the aim of this experiment was a comparison between the different configurations on a specific device.

Results show a large increase in both inference time and memory use when loading the on-premise solution due to the increased size of the model, proving the efficiency of our solution. More specifically, and by considering the layer conv4\_2 as a baseline, we experienced a 14.44\% inference time and 8.28\% memory usage increase in conv5\_1, 43.96\% inference time and 22.10\% memory usage increase in conv6\_1, 90.81\% inference time and 35.05\% memory usage increase in conv7, and 121.76\% inference time and 54.91\% memory usage increase in all layers (on premise). CPU usage also increases per configuration, however due to the multitasking nature of an android device, it is challenging to isolate the CPU usage of a single process and naturally the results fluctuates. Moreover, use of the lower intermediate layers can significantly reduce the complexity of private-feature extractors, especially when dealing with implementing complex deep architectures e.g. \emph{VGG-16} on edge devices and smartphones~\cite{kim2015compression}. 

Analyzing the complexity of different layers can lead us to considering accuracy-privacy-complexity trade-offs. As an example, consider Figure~\ref{fig:layer_comparison} and suppose we want to preserve the gender information. Comparing conv7 with conv4-2 and setting the accuracy to 95\%, we obtain 10\% more log-rank privacy with the cost of about 90\% more inference time. In this way we can choose the right strategy based on the importance of accuracy, privacy and complexity. Also by using the dimensionality reduction we can highly decrease the communication cost (compare the size of an image to size of 10 floating point numbers), although in this case we should consider the effect of dimensionality reduction on the complexity which is negligible.

We conclude that our algorithm can be implemented on a modern smartphone. By choosing a proper privacy-complexity trade-off and using different intermediate layers, we were able to significantly reduce the cost when running the model on a mobile device, while at the same time preserving important user information from being uploaded to the cloud.

\section{conclusion and future work}

In this paper, we proposed a hybrid framework for user data privacy preservation. This framework consists of a feature extractor and an analyzer module. The feature extractor provides a user with a private-feature which does not contains the user's desired sensitive information, but still maintains the required information to the service provider, so it can be used by the analyzer module in the cloud. In order to design the feature extractor, we used an information theoretic approach to formulate an optimization problem and proposed a novel deep architecture (\emph{DPFE}) to solve it. To measure the privacy of the extracted private-feature and verify the feature extractor, we proposed a new privacy measure called \emph{log-rank privacy}. Finally, we considered the problem of facial attribute prediction from face image, and attempted to extract a feature which contains facial attributes information while it does not contain identity information. By using \emph{DPFE} fine-tuning and implementing the model on mobile phone, we showed that we can achieve a reasonable tradeoff between facial attribute prediction accuracy, identity privacy and computational efficiency.

Our work can be extended in a number of ways. We used the proposed framework in an image processing application, while it can be used in other learning applications e.g. speech or text analysis and can be extended to other deep architectures e.g. recurrent neural networks. We formulated the problem for discrete sensitive variables but it can be extended for general cases. Analyzing the log-rank privacy measure can also have many potential applications in the privacy domain. An interesting future direction could be involving the log-rank privacy in the design of learning to rank algorithms. In an ongoing work, we are considering the challenge of privacy in a Machine Learning-as-a-Service platform.



%

\ifCLASSOPTIONcompsoc
  \section*{Acknowledgments}
\else
  \section*{Acknowledgment}
\fi

We acknowledge constructive feedback from Sina Sajadmanesh, Amirhossein Nazem and David Meyer. Hamed Haddadi was supported by the EPSRC Databox grant (Ref: EP/N028260/1), EPSRC IoT-in-the-Wild grant (Ref: EP/L023504/1), and a Microsoft Azure for Research grant. 


\ifCLASSOPTIONcaptionsoff
  \newpage
\fi



\bibliographystyle{IEEEtran}
\bibliography{IEEEabrv,ref2}

\appendices
\section{Preliminaries}\label{appendix:A}

Quantizing some intuitive concepts like uncertainty and information, is one of the main information theory advantages. In this part we briefly discuss these phenomenons and refer the readers to further detailed discussion in~\cite{cover2012} and~\cite{haykin2009}.

The \emph{entropy} of a discrete random variable $\textbf{x}$ is defined as:
\begin{align*}
H(\textbf{x}) = \mathds{E}_\textbf{x}  [-\log p(x)]
\end{align*} 
which can be used to measure the uncertainty we have about $\textbf{x}$. \emph{Differential entropy} is the extention of this definition for the continuous random variables: $h(\textbf{x}) = -\int f(x) \log f(x) dx$, where here $f(x)$ is the probability density function of $\textbf{x}$. We can also define entropy for joint and conditional probability distributions:
\begin{align*}
H(\textbf{x},\textbf{y}) &= \mathds{E}_{\textbf{x},\textbf{y}}  [-\log p(x,y)] \\
H(\textbf{x}|\textbf{y}) &= \mathds{E}_{\textbf{y}} \mathds{E}_{\textbf{x}|\textbf{y}}  [-\log p(x|y)] 
\end{align*}

Based on these definitions, we can define the \emph{mutual information} between two random variables, which tries to measure the amount of uncertainty reduction about one of them, given the other one:
\begin{align*}
I(\textbf{x};\textbf{y}) = H(\textbf{x}) - H(\textbf{x}|\textbf{y}) = H(\textbf{y}) - H(\textbf{y}|\textbf{x})
\end{align*}
It is also equal to $kl$-divergence between $p(x,y)$ and $p(x)p(y)$. $kl$-divergence between two probability distributions $p$ and $q$ is a non-negative distance measure between them, define as:
\begin{align*}
D_{kl}[p\|q] = \mathds{E}_p [\log \frac{p}{q}]
\end{align*}
So we have $I(\textbf{x};\textbf{y}) = D_{kl}[p(x,y)\|p(x)p(y)]$. These are the information theoretic definitions we used to define and solve the privacy preservation problem. Further information can be accessed through~\cite{cover2012}.

\section{}\label{appendix:B}

\subsection{Proof of Lemma \ref{lem0}}
\label{appendix:B_1}
From positivity of $kl$-divergence we know that:
\begin{align*}
kl(p(z|f)\|q(z|f)) = \int p(z|f) \log \frac{p(z|f)}{q(z|f)} \; dz 	\geq 0
\end{align*}
So we have:
\begin{align*}
\int p(f,z) \log \frac{p(z|f)\:p(z)}{p(z)\: q(z|f)} \; dz \; df 	\geq 0
\end{align*}
Also we know that:
\begin{align*}
&I(\textbf{f};\textbf{z}) = \int p(f,z) \log \frac{p(f,z)}{p(f)p(z)} \; df \;dz \\
 						 &= \int p(z) \int p(f|z) \log \frac{p(z|f)}{p(z)}\; df \; dz					 
\end{align*}
Thus:
\begin{align*}
I(\textbf{f};\textbf{z}) \geq 
\int p(f,z) \log \frac{q(z|f)}{p(z)} \; dz \; df 	
\end{align*}

\subsection{Proof of Theorem \ref{th1}}
\label{appendix:B_2}
From Lemma~\ref{lem1} we know that:
\begin{align*}
\begin{split}
&I(\textbf{f};\textbf{y}) = \sum_a p(y_a) \int p(f|y_a) \log \frac{p(f|y_a)}{\sum_b p(y_b)p(f|y_b)}\ df \\
&= -\sum_a p(y_a) \bigg[ H(\textbf{f}|y_a) + \int p(f|y_a) \log\mathds{E}_{\textbf{y}} p(f|y)df  \bigg]
\end{split}
\end{align*} 
So by using Jensen inequality we have:
\begin{align*}
\begin{split}
I(\textbf{f};\textbf{y}) \leq & -\sum_a p(y_a) \bigg[ H(\textbf{f}|y_a) + \int p(f|y_a) \mathds{E}_{\textbf{y}}\log p(f|y)df  \bigg]\\
& = \mathcal{U}_1
\end{split}
\end{align*}
We can manipulate $\mathcal{U}_1$ as:
\begin{align*}
\begin{split}
\mathcal{U}_1 = \sum_a p(y_a) \int p(f|y_a) \Big[\log p(f|y_a)\\
 -   \sum_b p(y_b) \log p(f|y_b)df \Big]
\end{split}
\end{align*}
So we get:
\begin{align*}
\begin{split}
\mathcal{U}_1 &= \sum_a \sum_b p(y_a)\ p(y_b)\ D_{kl} \big[ p(f|y_a) \| p(f|y_b) \big] \\
&= \sum_a \sum_{b: b\neq a} p(y_a)\ p(y_b)\ D_{kl} \big[ p(f|y_a) \| p(f|y_b) \big] 
\end{split}
\end{align*} 


\subsection{Proof of Theorem \ref{th2}}
\label{appendix:B_3}
By using Lemma~\ref{lem2} and \ref{lem3} we get:
\begin{align}
\begin{split}
&\mathcal{U}_1 \simeq \sum_a \sum_{b:\ b\neq a} \frac{N_{y_a}}{N}\ \frac{N_{y_b}}{N}\ D_{kl} \big[ p(f|y_a) \| p(f|y_b) \big] \\
&\leq  \sum_a \sum_{b:\ b\neq a} \frac{N_{y_a}}{N}\ \frac{N_{y_b}}{N}\ \sum_{\substack{(i,j):\\y_i=y_a\\y_j=y_b} }  \frac{1}{N_{y_a}}\ \frac{1}{N_{y_b}} \frac{1}{2\sigma}\ \|f_i-f_j\|_2^2 \\
&= \frac{1}{\sigma N^2} \sum_{(i,j):\ y_i\neq y_j} \|f_i-f_j\|_2^2 = \mathcal{U}_2
\end{split}
\end{align}

\subsection{Proof of Theorem \ref{th3}}
\label{appendix:B_4}
In order to prove this theorem, first we need to address the following lemma:
\begin{lemma}
	Assuming $f_1$ and $f_2$ are two samples from $p(f)$ with mean $\mu$ and covariance matrix $\Sigma$ we have:
	\begin{align*}
	\begin{split}
	\mathds{E} \Big[ \|f_1 - f_2\|_2^2 \Big] &= \mathds{E} \Big[(f_1-f_2)^T(f_1-f_2)\Big]\\ &= 2 \text{diag}(\Sigma + \mu \mu^T) - 2\mu^T \mu = 2 \text{diag}\Sigma
	\end{split}
	\end{align*}
\end{lemma}
So by normalizing the feature space to has variance one for each dimension, $\mathds{E} \Big[ \|f_1 - f_2\|_2^2 \Big]$ is fixed and equal to $2d$ where $d$ is the dimension.

Now we can state the proof of Theorem~\ref{th3}. Considering $\{f_i\}_{i=1}^N$ as $i.i.d.$ samples from $p(f)$ and setting $d_{ij} = \|f_i - f_j\|_2^2$ we have:
\begin{align*}
\sum_{i,j} d_{ij} \simeq \binom{N}{2} 2d
\end{align*}
We can also split pairs with their $\textbf{y}$ labels similarity:
\begin{align*}
\sum_{i,j: y_i=y_j} d_{ij} + \sum_{i,j: y_i\neq y_j} d_{ij}  \simeq \binom{N}{2} 2d
\end{align*}
and get:
\begin{align*}
\begin{split}
\frac{1}{N^2}\sum_{i,j: y_i\neq y_j} d_{ij}   \simeq &\frac{2d(N-1)}{N} - \frac{1}{N^2}\sum_{i,j: y_i= y_j} d_{ij}\\
& = \frac{1}{N^2} \sum_{i,j: y_i= y_j}\big( \frac{2dN(N-1)}{k} - d_{ij} \big)
\end{split}
\end{align*}
where $k$ is the number of similar pairs in the training data.

%

%

\begin{IEEEbiography}[{\includegraphics[width=1in,height=1.25in,clip,keepaspectratio]{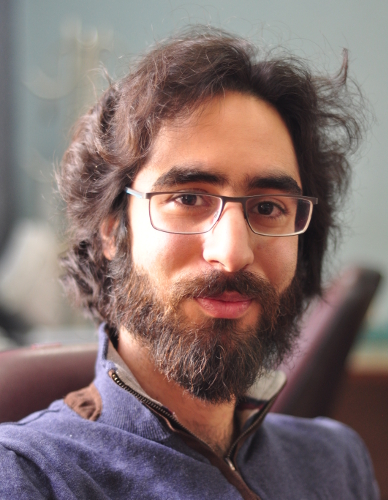}}]{Seyed Ali Osia}
	received his B.Sc. degree in Software Engineering from Sharif University of Technology in 2014. He is currently a Ph.D. candidate at the department of computer engineering, Sharif University of Technology. His research interests includes Statistical Machine Learning, Deep Learning, Privacy and Computer Vision. 
\end{IEEEbiography}
\begin{IEEEbiography}[{\includegraphics[width=1in,height=1.25in,clip,keepaspectratio]{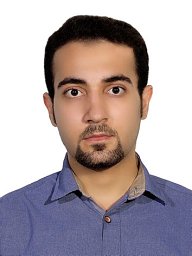}}]{Ali Taheri}
	received his B.Sc. degree in Software Engineering from Shahid Beheshti University in 2015. He received his M.Sc. degree in Artificial Intelligence from Sharif University of Technology in 2017. His research interests includes Deep Learning and Privacy.
\end{IEEEbiography}
\begin{IEEEbiography}[{\includegraphics[width=1in,height=1.25in,clip,keepaspectratio]{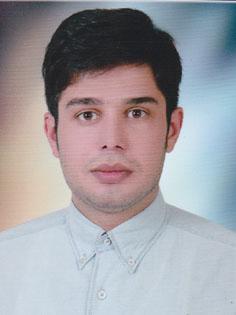}}]{Ali Shahin Shamsabadi}
	received his B.S. degree in electrical engineering from Shiraz University of Technology, in 2014, and the M.Sc. degree in electrical engineering (digital) from the Sharif University of Technology, in 2016. Currently, he is a Ph.D. candidate at the Queen Mary University of London. His research interests include deep learning and data privacy protection in distributed and centralized learning. 
\end{IEEEbiography}
\begin{IEEEbiography}[{\includegraphics[width=1in,height=1.25in,clip,keepaspectratio]{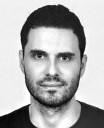}}]{Kleomenis~Katevas}
	received his B.Sc. degree in Informatics Engineering from the University of Applied Sciences of Thessaloniki in 2006, and an M.Sc. degree in Software Engineering from Queen Mary University of London in 2010. He is currently a Ph.D. candidate at Queen Mary University of London. His research interests includes Mobile \& Ubiquitous Computing, Applied Machine Learning, Crowd Sensing and Human-Computer Interaction.
\end{IEEEbiography}

\begin{IEEEbiography}[{\includegraphics[width=1in,height=1.25in,clip,keepaspectratio]{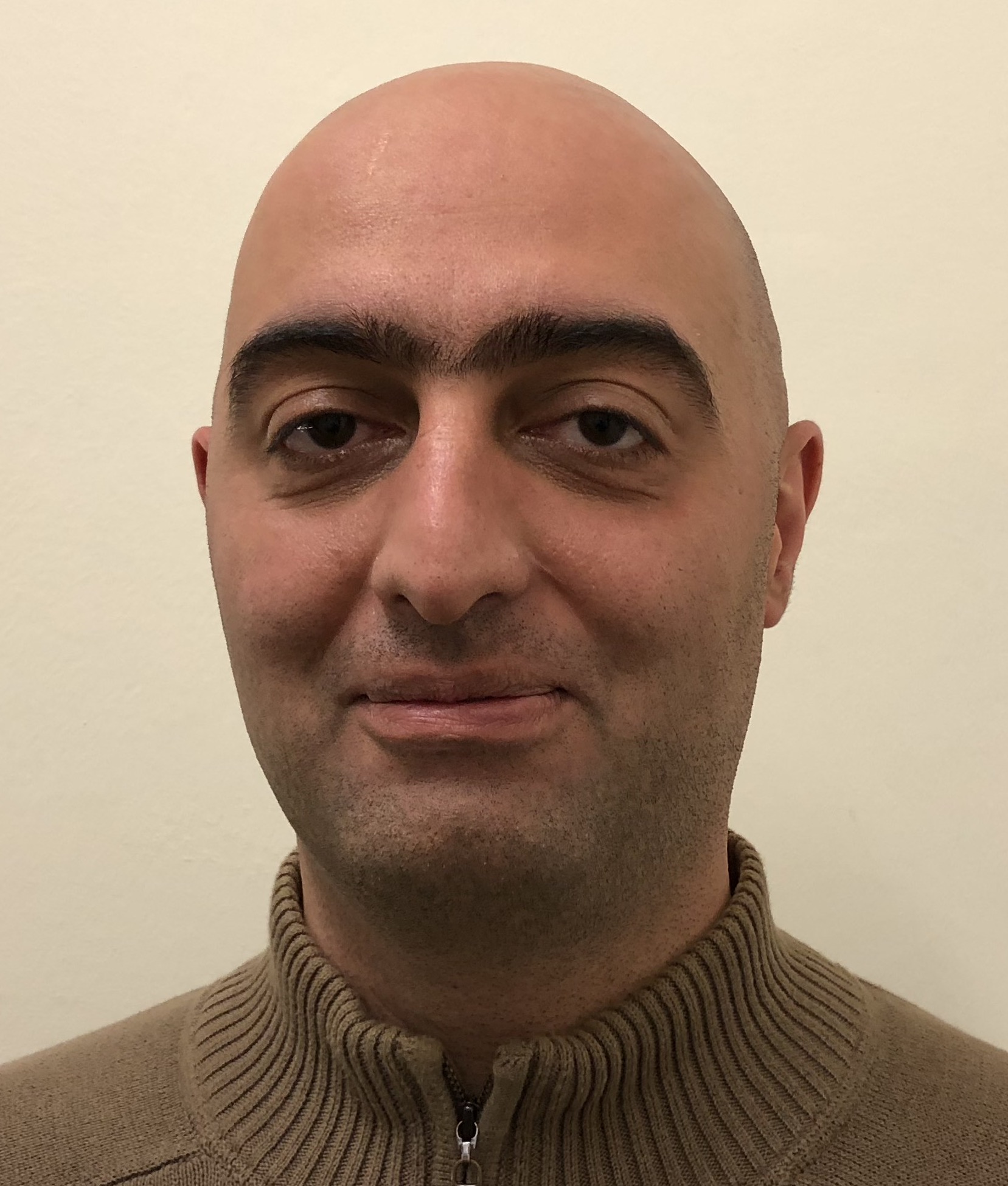}}]{Hamed Haddadi}
received his B.Eng., M.Sc., and Ph.D. degrees from University College London. He was a postdoctoral researcher at Max Planck Institute for Software Systems in Germany, and a postdoctoral research fellow at Department of Pharmacology, University of Cambridge and The Royal Veterinary College, University of London, followed by few years as a Lecturer and consequently Senior Lecturer in Digital Media at Queen Mary University of London. He is currently a Senior Lecturer (Associate Professor) and the Deputy Director of Research in the Dyson School of Design Engineering, and an Academic Fellow of the Data Science Institute, at The Faculty of Engineering at Imperial College of London. He is interested in User-Centered Systems, IoT, Applied Machine Learning, and Data Security \& Privacy. He enjoys designing and building systems that enable better use of our digital footprint, while respecting users' privacy. He is also broadly interested in sensing applications and Human-Data Interaction. 
\end{IEEEbiography}

\begin{IEEEbiography}[{\includegraphics[width=1in,height=1.25in,clip,keepaspectratio]{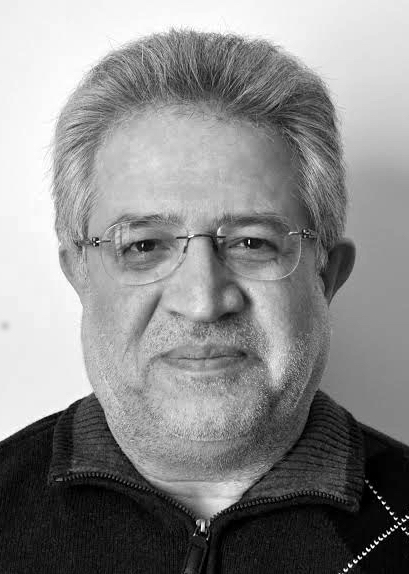}}]{Hamid R. Rabiee}
received his B.S. and M.S. degrees (with great distinction) in electrical engineering from California State University, Long Beach, CA, in 1987 and 1989, respectively; the EEE degree in electrical and computer engineering from the University of Southern California (USC), Los Angeles, CA; and the Ph.D. degree in electrical and computer engineering from Purdue University, West Lafayette, IN, in 1996. From 1993 to 1996, he was a Member of the Technical Staff at AT\&T Bell Laboratories. From 1996 to 1999, he worked as a Senior Software Engineer at Intel Corporation. From 1996 to 2000, he was an Adjunct Professor of electrical and computer engineering with Portland State University, Portland, OR; with Oregon Graduate Institute, Beaverton, OR; and with Oregon State University, Corvallis, OR. Since September 2000, he has been with the Department of Computer Engineering, Sharif University of Technology, Tehran, Iran, where he is a Professor of computer engineering, and Director of Sharif University Advanced Information and Communication Technology Research Institute (AICT), Digital Media Laboratory (DML), and Mobile Value Added Services Laboratory (MVASL). He is also the founder of AICT, Advanced Technologies Incubator (SATI), DML, and VASL. He is currently on sabbatical leave (2017-2018 academic year) as visiting professor at Imperial College of London. He has been the Initiator and Director of national and international-level projects in the context of United Nation Open Source Network program and Iran National ICT Development Plan. He has received numerous awards and honors for his industrial, scientific, and academic contributions. He is a Senior Member of IEEE, and holds three patents. He has also initiated a number of successful start-up companies in cloud computing, SDP, IoT, and storage systems for big data analytics. His research interests include statistical machine learning, Bayesian statistics, data analytics and complex networks with applications in multimedia systems, social networks, cloud and IoT data privacy, bioinformatics, and brain networks.
\end{IEEEbiography}






\end{document}